%% file: Pedestrian_TPAMI.tex
\documentclass[lettersize,journal]{IEEEtran}
\usepackage{amsmath,amsfonts}
\usepackage{algorithmic}
\usepackage{algorithm}
\usepackage{array}
\usepackage[caption=false,font=normalsize,labelfont=sf,textfont=sf]{subfig}
\usepackage{textcomp}
\usepackage{stfloats}
\usepackage{url}
\usepackage{verbatim}
\usepackage{graphicx}
\usepackage{cite}

\usepackage{siunitx}
\usepackage{multirow}
\usepackage{makecell}
\usepackage{textcomp}
\usepackage{stfloats}
\usepackage{url}
\usepackage{verbatim}
\usepackage{balance}
\usepackage{orcidlink}
\usepackage{array}
\usepackage{booktabs}
\usepackage{threeparttable}
\usepackage{bm}

\begin{document}
\title{PedestrianDiffusion: Multimodal Generative Denoising and Dense State Estimation for Inertial Navigation}
\author{
    I-Hao Lu\,\orcidlink{0009-0000-1167-7488}, Dongsoo Han\,\orcidlink{0000-0001-9844-4221} ~\IEEEmembership{Senior Member,~IEEE,}
    \thanks{
        The source code for the proposed model is available on GitHub at \url{https://github.com/jacklu333333/PedestrianDiffusion}
    }
}
\markboth{Journal of \LaTeX\ Class Files,~Vol.~14, No.~8, August~2021}%
{Shell \MakeLowercase{\textit{et al.}}: A Sample Article Using IEEEtran.cls for IEEE Journals}


\maketitle
\begin{abstract}
    The accuracy of consumer-grade inertial navigation is bottlenecked by the stochastic noise of Micro-Electro-Mechanical Systems (MEMS). Traditional deterministic neural architectures often succumb to ``estimation jittering,'' sacrificing high-frequency kinematic fidelity for numerical stability. We propose PedestrianDiffusion, a multimodal spectral-domain generative framework reformulating dense 6D state estimation as a continuous conditional denoising process. By operating in the frequency domain, our formulation bounds the spectral covariance, acting as a mathematical preconditioner to stabilize the reverse diffusion trajectory. Furthermore, we introduce a zero-shot semantic conditioning mechanism leveraging vision-language embeddings as categorical priors to generalize across heterogeneous sensor noise profiles. To address the computational intractability of generative tracking, we deploy a single-step deterministic probability flow ODE solver ($T=1$). This yields high-capacity asynchronous batch trajectory refinement, establishing the viability of generative architectures for asynchronous batch trajectory refinement on edge hardware. Extensive evaluations on the OxIOD, RIDI, RoNIN, and TLIO benchmarks demonstrate that PedestrianDiffusion achieves state-of-the-art performance, exhibiting unprecedented robustness to impulse perturbations and coupled 6D kinematic drift. This work provides a rigorous algorithmic blueprint for next-generation Neural Inertial Measurement Units (N-IMUs).
\end{abstract}


\section{Introduction}

Despite the widespread integration of low-cost MEMS Inertial Measurement Units (IMUs), achieving navigation-grade accuracy remains elusive due to stochastic noise and bias instability~\cite{9195512}. Traditional filtering techniques (e.g., ZUPT~\cite{7102742}, EKF~\cite{9941084}) depend on restrictive physical priors that frequently fail in unconstrained environments. While deep learning architectures such as RoNIN~\cite{9196860} and TLIO~\cite{liu2020tlio} provide robust alternatives, they predominantly formulate navigation as a regression task. However, this paradigm has encountered a ``estimation jittering'' caused by optimizing for Mean Squared Error (MSE), discriminative models inevitably overestimate the velocity and lose high-frequency kinematic details to minimize variance. Consequently, these models yield jittered, overestimated trajectories---statistically reflected by Total Length Ratios (TLR) consistently below unity---trading physical fidelity for numerical stability.

To overcome this limitation, we propose a conditional generative denoising paradigm that fundamentally departs from discriminative regression. We introduce PedestrianDiffusion (PD), a multimodal spectral-domain diffusion framework designed to actively reconstruct the clean motion manifold from noisy sensor inputs. By shifting from time-domain regression to spectral processing via the Short-Time Fourier Transform (STFT), our approach inherently isolates periodic pedestrian kinematics from stochastic sensor noise. This formulation enables PD to achieve optimal accuracy across the OxIOD~\cite{Chen_Lu_Markham_Trigoni_2018}, RIDI~\cite{yan2017ridirobustimudouble}, RoNIN~\cite{9196860}, and TLIO~\cite{liu2020tlio} benchmarks.

Furthermore, we challenge the prevailing consensus that generative models are computationally prohibitive for edge deployment. Our empirical analysis yields a critical insight: spectral diffusion exhibits rapid convergence, saturating state estimation accuracy in a single inference step ($T=1$). This convergence translates to an amortized latency of \qty{9.88}{ms} per frame per inference. While strict hard real-time execution currently necessitates discrete hardware accelerators due to sequential dependencies, this amortized efficiency proves that generative tracking is computationally viable for high-throughput edge buffering. Ultimately, PD serves as an algorithmic proof-of-concept for next-generation Neural Inertial Measurement Units (N-IMUs), demonstrating that the computational demands of generative denoising constitute a justified investment for tactical-grade precision.

The primary contributions of this paper are:
\begin{itemize}
    \item \textbf{Generative Navigation Paradigm:} We pioneer a spectral diffusion framework for inertial navigation, transitioning the field from MSE-based regression to high-fidelity generative modeling capable of recovering high-frequency kinematic details.
    \item \textbf{Spectral \& Physical Consistency:} We demonstrate that processing IMU data in the frequency domain preserves superior signal resolution compared to temporal methods. We enforce kinematic coherence via a novel dual-domain loss function that jointly optimizes spectral reconstruction and rigid-body dynamics.
    \item \textbf{Device-Agnostic Adaptation:} We propose a CLIP-based conditioning mechanism that adapts to heterogeneous device noise profiles (e.g., ``handheld'' vs. ``handheld speed''). By leveraging semantic proximity, this strategy enables zero-shot categorical clustering to address the unseen sensor problem without model retraining.
\end{itemize}
\section{Related Work}
\label{sec:related_work}

\noindent \textbf{Classical Methods: The Sensor Placement Barrier.}
Classical inertial navigation systems traditionally employ Extended Kalman Filters (EKFs) constrained by physical priors to bound error accumulation. A foundational approach is the Zero-Velocity Update (ZUPT)~\cite{sher1996personal}, which mitigates velocity drift by detecting stationary phases during the gait cycle. Extending this principle, NavShoe~\cite{1528431} integrated ZUPT within a filtering framework to arrest cubic error growth, achieving a terminal position error of \qty{0.34}{m} over \qty{118.5}{m}. Subsequently, Skog et al.~\cite{5646936} provided a rigorous evaluation of the zero-velocity detectors requisite for enforcing these constraints.

However, the efficacy of classical methods is strictly bottlenecked by sensor placement. Their reliance on discernible stationary periods (e.g., foot-mounted IMUs) renders them fundamentally unsuitable for mass-market consumer electronics, such as smartphones or VR headsets, which are characterized by continuous motion and variable lever arms. Consequently, accurate navigation remains heavily constrained on ubiquitous devices where zero-velocity assumptions cannot be reliably satisfied.

\noindent \textbf{Deep Learning Methods: Estimation Jittering.}
Deep inertial odometry circumvents placement constraints by formulating state estimation as data-driven sequence modeling. Building upon IONet~\cite{Chen_Lu_Markham_Trigoni_2018} and RoNIN~\cite{9196860}, CTIN~\cite{rao2022ctin} advanced the paradigm via a Robust Contextual Transformer. This architecture couples a ResNet encoder with local and global self-attention to capture spatial dependencies, utilizing a Transformer decoder for temporal fusion and multi-task learning to jointly estimate velocity and uncertainty covariance.

Despite these architectural advancements, current discriminative models are bounded by a ``estimation jittering'' in velocity. The pervasive reliance on MSE optimization inevitably overestimates details to minimize variance. This optimization objective yields blurred, drifting trajectories---statistically evidenced by TLR consistently deviating from unity---sacrificing the physical fidelity requisite for high-precision navigation.

\noindent \textbf{Hybrid Methods: Sampling Efficiency Gap.}
Hybrid architectures seek to reconcile the representational capacity of deep learning with the theoretical guarantees of recursive filtering. For instance, TLIO~\cite{liu2020tlio} proposed a tightly-coupled framework wherein a neural network acts as a pseudo-measurement generator for an EKF. This synergy facilitates drift correction in the absence of strictly stationary phases, surpassing pure learning baselines. LLIO~\cite{wang2022llio} later addressed similar objectives and claimed a more lightweight formulation.

A primary limitation of hybrid systems like TLIO~\cite{liu2020tlio} is their strict dependence on high-frequency sampling. Maintaining EKF stability frequently demands IMU sampling rates of approximately \qty{1000}{Hz}~\cite{liu2020tlio}, a specification that heavily penalizes the power and bandwidth constraints of commodity devices. In contrast, our generative approach achieves superior estimation accuracy at a standard \qty{100}{Hz} sampling rate, demonstrating that expressive sequence modeling can effectively compensate for reduced temporal resolution.

\noindent \textbf{Coordinate Normalization via Learned Canonical Frames.}
To address the sensitivity of learning-based odometry to sensor orientation, EqNIO~\cite{jayanth2025eqnio} learns an equivariant canonical frame directly from inertial measurements. By mapping IMU signals into a gravity-aligned coordinate system, this method enables downstream networks to process orientation-invariant representations. Jointly optimizing this canonicalization with the odometry objective significantly improves zero-shot generalization across diverse sensor orientations and motion trajectories.

\noindent \textbf{Generative Navigation: From Time to Spectral Domain.}
The unprecedented success of diffusion models in continuous signal synthesis has recently motivated their application to trajectory estimation. While Yu et al.~\cite{yu2024trajectory} validated diffusion for multimodal object-goal planning, DiffusionIMU~\cite{teng2025diffusionimu} introduced the architecture to inertial navigation by denoising velocity estimates in the time domain.

Unlike DiffusionIMU's reliance on time-domain regression, PedestrianDiffusion formulates state estimation strictly in the spectral domain via the Short-Time Fourier Transform (STFT). By explicitly modeling the periodic structure of pedestrian kinematics, we mathematically isolate high-frequency details from stochastic noise. This spectral formulation not only maximizes physical reconstruction fidelity but critically enables the rapid convergence ($T=1$) necessary to render low-latency generative navigation computationally tractable on edge devices.

\section{Methodology}
\subsection{Spectral Disparity in Linear Sensor Spaces}
\label{sec:linear_diffusion_instability}

While diffusion models excel in bounded domains like natural images \cite{ho2020denoising, 10.5555/3540261.3540933, ramesh2022hierarchical}, applying standard formulations to unbounded inertial data introduces instability. This failure arises from a fundamental mismatch between the isotropic Gaussian prior used in diffusion and the highly anisotropic spectral structure of inertial signals.

\paragraph{The Isotropic Assumption.}
Let $x_0 \in \mathbb{R}^d$ denote the clean signal. The standard forward process \cite{ho2020denoising, sohl2015deep}, defined as
\begin{equation}
    x_t = \sqrt{\alpha_t} x_0 + \sqrt{1 - \alpha_t}\,\epsilon, \quad \epsilon \sim \mathcal{N}(0, I),
    \label{eq:forward_diffusion}
\end{equation}
implicitly imposes an isotropic noise geometry. This is appropriate for image data, where pixel values are bounded (e.g., $[0,1]$), and energy is relatively evenly distributed across spatial dimensions. However, inertial measurements are unbounded and exhibit extreme spectral skew: low-frequency integration drift often carries variance magnitudes higher than high-frequency motion dynamics.

\paragraph{Ill-Conditioned Reverse Dynamics.}
The incompatibility becomes evident in the signal covariance. For a signal with covariance $\Sigma_0 = \mathrm{Cov}(x_0)$, the perturbed covariance is given by:
\begin{equation}
    \Sigma_t = \alpha_t \Sigma_0 + (1 - \alpha_t) I.
\end{equation}
In inertial spaces, $\Sigma_0$ is typically rank-deficient or ill-conditioned due to physical constraints and sensor biases. As $t \to 1$, the isotropic term $(1-\alpha_t)I$ overwhelms the signal structure in directions corresponding to small eigenvalues of $\Sigma_0$. Conversely, directions with large eigenvalues (e.g., drift) require excessive noise levels to be effectively diffused.

This disparity leads to a heteroscedastic signal-to-noise ratio (SNR) across dimensions. During the reverse process, the score network must approximate $\nabla_{x_t} \log p(x_t)$ \cite{song2021scorebasedgenerativemodelingstochastic}, effectively inverting this noise. For dimensions where the SNR is vanishingly small, the denoising task becomes ill-posed, resulting in numerical divergence—observed empirically as value explosion.

\paragraph{Spectral Conditioning.}
To remedy this, we propose aligning the diffusion process with the intrinsic geometry of the sensor space. By operating on a frequency-aligned basis and scaling noise injection according to the signal's spectral energy, we ensure a balanced SNR across all modes. This geometric alignment stabilizes training and, as shown in our experiments, allows for saturation of performance at significantly lower parameter counts by removing the burden of learning to compensate for spectral mismatch.

\subsection{Spectral-Domain Diffusion for Unbounded Signals}
\label{sec:spectral_diffusion}
To decouple the denoising objective from global orientation, we align input sequences to a gravity-stabilized, heading-agnostic coordinate frame (HACF)~\cite{9196860}. We achieve the HACF by performing North-East rotation data augmentation. We then employ the Short-Time Fourier Transform (STFT) to mitigate the dynamic range issue and to separate periodic kinematic structure from high-frequency sensor noise. Processing $100$\,Hz inertial signals over one-second intervals, this transformation yields a compact $(16, 8, 2)$ time--frequency representation that preserves physically meaningful amplitude information while avoiding the failure modes of naive scalar normalization. The detailed spectral configuration, including windowing and Fourier parameters, is provided in \autoref{sec:appendix_input}.

\subsection{Global Standardization Strategy}
\label{sec:global_standardization}
To preserve physical scale consistency during diffusion, we avoid instance-level normalization and instead adopt a global standardization strategy. Specifically, we compute a dataset-wide scaling factor based on the maximum empirical standard deviation observed across all training samples, ensuring that the majority of signals lie within a numerically stable range while maintaining relative amplitude relationships. Implementation details and normalization hyperparameters are reported in \autoref{sec:appendix_input}.

\subsection{PedestrianDiffusion Model}
\input{figure/Diffusion_overview}
\autoref{fig:Diffusion_overview} illustrates the PD architecture, which integrates inpainting~\cite{lugmayr2022repaint,lin2024joint} with multimodal conditioning~\cite{mao2023leapfrog} for robust generation. The framework comprises a Variational Autoencoder (VAE), a text encoder, and a diffusion-based denoising network. Input IMU data is STFT-transformed and normalized (scaled $5\times$ relative to \autoref{tab:Normalization}) to maximize dynamic range coverage, then concatenated with noisy state embeddings. A UNet, conditioned on fused VAE latents and CLIP embeddings, iteratively reconstructs high-fidelity trajectories.

\input{figure/sentence}
\noindent \textbf{Variational Autoencoder \& Text Encoding.}
To learn a compact and noise-resilient representation of pedestrian motion, we employ a VAE~\cite{kingma2013auto} as a spectral feature extractor. The VAE is trained to encode input frequency-domain IMU signals into a latent variable $\mathbf{z} \in \mathbb{R}^{512 \times 1}$. This projection encapsulates essential motion primitives while attenuating high-frequency sensor noise, thereby providing a structured feature space for the subsequent diffusion process. By leveraging the VAE's generative capabilities, we enhance the model's ability to generalize across diverse motion patterns.

To further improve robustness against distribution shifts---a common challenge when training and testing on disjoint domains---we incorporate auxiliary metadata provided by datasets such as RIDI~\cite{yan2017ridirobustimudouble} and OxIOD~\cite{chen2018oxioddatasetdeepinertial}. As illustrated in \autoref{fig:sentence}, we serialize discrete attributes (e.g., device model, mounting position) into natural language prompts. We then utilize the CLIP ViT-B/32 model~\cite{radford2021learning} to encode these prompts into semantic text embeddings $\mathbf{c} \in \mathbb{R}^{512 \times 77}$. While the pre-trained CLIP model inherently possesses no physical or kinematic comprehension of inertial MEMS sensor noise, its text embeddings serve an invaluable epistemological role in our framework. By mapping textual metadata (e.g., device placement, model type) into high-dimensional semantic spaces, CLIP acts as a sophisticated, zero-shot categorical clustering and hashing mechanism. It constructs distinct decision boundaries that map specific text descriptors to highly correlated noise distributions observed during training. This categorical conditioning stabilizes denoising performance by allowing the network to explicitly associate specific sensor noise profiles with specific semantic clusters, facilitating robust generalization to unseen sensors via semantic proximity rather than physical reasoning.

To prevent overfitting to fixed syntactic patterns, we iterate through various sentence permutations during training, as shown in \autoref{fig:sentence}. Furthermore, we introduce an ``unknown'' category---computed as the average embedding of all user profiles---which allows the model to generalize to uncharacterized configurations during inference.

Finally, we synthesize a unified input tensor by concatenating the motion latent $\mathbf{z}$ and the contextual embedding $\mathbf{c}$ along the feature dimension. The resulting representation $\mathbf{x} \in \mathbb{R}^{512 \times 78}$ is fed into the diffusion network, enabling it to jointly leverage learned spectral features and semantic context for precise trajectory reconstruction.

\noindent \textbf{Model Architecture.}
While the Transformer-based DiT~\cite{Peebles2022DiT} excels at scale, its lack of inductive bias is detrimental in the data-scarce inertial domain. In contrast, the convolutional UNet~\cite{xie2023diffusionmodelgenerativeimage} provides essential priors for local temporal dynamics. Our ablations confirm that these priors confer superior stability and sample efficiency; accordingly, we adopt a conditional 3D UNet backbone.

\subsection{Diffusion Backbone and Inference}
\label{sec:scheduler}
We adopt the DPM-Solver++ framework~\cite{lu2025dpm}, utilizing the Squared Cosine Schedule (v2)~\cite{nichol2021improved} to prevent boundary singularities. With $T=1000$, the cumulative retention $\bar{\alpha}_t$ is defined as:
\begin{equation}
    \bar{\alpha}_t = \frac{f(t)}{f(0)}, \quad f(t) = \cos^2 \left( \frac{t/T + s}{1 + s} \cdot \frac{\pi}{2} \right),
    \label{eq:cosine_schedule}
\end{equation}
where $s$ is a small offset. The model $f_\theta(x_t, t)$ is configured to predict the clean data $x_0$ directly by minimizing the reconstruction error:
\begin{equation}
    \mathcal{L} = \mathbb{E}_{x_0, t, \epsilon} \left[ \| x_0 - f_\theta(\sqrt{\bar{\alpha}_t} x_0 + \sqrt{1 - \bar{\alpha}_t} \epsilon, t) \|^2 \right].
    \label{eq:objective}
\end{equation}

\begin{figure}
    \centering
    \begin{minipage}{0.48\textwidth}
        \begin{algorithm}[H]
            \caption{Training}
            \label{alg:training}
            \begin{algorithmic}[1]
                \REPEAT
                \STATE Sample $x_0 \sim q(x_0)$, $\epsilon \sim \mathcal{N}(\mathbf{0}, \mathbf{I})$
                \STATE Sample $t \sim \text{Uniform}(\{1, \ldots, T\})$
                \STATE $\bar{\alpha}_t \leftarrow$ Eq.~\ref{eq:cosine_schedule}
                \STATE $x_t = \sqrt{\bar{\alpha}_t} x_0 + \sqrt{1 - \bar{\alpha}_t} \epsilon$
                \STATE $\theta \leftarrow \theta - \eta \nabla_\theta \| x_0 - f_\theta(x_t, t) \|^2$
                \UNTIL converged
            \end{algorithmic}
        \end{algorithm}
    \end{minipage}
    \hfill
    \begin{minipage}{0.48\textwidth}
        \begin{algorithm}[H]
            \caption{Simplified DPM-Solver++ }
            \label{alg:sampling_2m_simple}
            \begin{algorithmic}[1]
                \STATE Sample $x_T \sim \mathcal{N}(\mathbf{0}, \mathbf{I})$
                \FOR{$t = T, \ldots, 1$}
                \STATE $\hat{x}_0^{(t)} = f_\theta(x_t, t)$
                \STATE $h_t = \lambda_{t-1} - \lambda_t$

                \IF{$t = T$}
                \STATE $\tilde{x}_0 = \hat{x}_0^{(t)}$
                \ELSE
                \STATE $r_t = h_t / h_{t+1}$
                \STATE $\tilde{x}_0 = \hat{x}_0^{(t)} + \frac{1}{2} r_t \left( \hat{x}_0^{(t)} - \hat{x}_0^{(t+1)} \right)$
                \ENDIF

                \STATE $c_1 = \frac{\sqrt{1 - \bar{\alpha}_{t-1}}}{\sqrt{1 - \bar{\alpha}_t}} e^{-h_t}$
                \STATE $c_2 = \sqrt{1 - \bar{\alpha}_{t-1}} (1 - e^{-h_t})$
                \STATE $x_{t-1} = c_1 x_t + c_2 \tilde{x}_0$
                \ENDFOR
                \STATE \textbf{return} $x_0$
            \end{algorithmic}
        \end{algorithm}
    \end{minipage}
\end{figure}

\noindent \textbf{DPM-Solver Scheduler++.} For efficient inference, we utilize the DPM-Solver++~\cite{lu2022dpmsolver} to solve the diffusion Probability Flow ODE. We employ the second-order deterministic solver (Order 2) to eliminate ancestral stochasticity. Furthermore, we enforce Zero-SNR rescaling to ensure the signal-to-noise ratio approaches zero at $T$, reducing exposure bias.
\subsection{Dual-Domain Consistent Loss Function}
Since we have dual objectives of denoising and high-fidelity estimation for both velocity and angular velocity, we design a dual-domain loss function by referencing the work of IoNet~\cite{Chen_Lu_Markham_Trigoni_2018,kendall2018multi}.
\input{equation/loss_function}

\subsection{Single-Step Inference Theory}
Recent advancements \cite{song2023consistency,yin2024one} establish a rigorous theoretical foundation for accelerated sampling in diffusion models. By reformulating the iterative generation process as either a self-consistency mapping along the probability flow ordinary differential equation (PF-ODE) trajectory \cite{song2023consistency} or a direct distribution matching task \cite{yin2024one}, these frameworks circumvent the cumulative discretization errors inherent in standard numerical solvers. Theoretically, they demonstrate that the statistical distance $D$ between the generated distribution $p_\theta$ and the true data distribution $p_{data}$ is strictly bounded by the empirical optimization objective $\mathcal{L}(\theta)$:
$$ D(p_{data} \| p_\theta) \leq \mathcal{O}(\mathcal{L}(\theta) + \epsilon) $$
where $\epsilon$ denotes the irreducible approximation error. This theoretical guarantee ensures that the inference budget can be reduced to a minimal Number of Function Evaluations (NFEs) without catastrophic degradation in sample fidelity.

\section{Evaluation}
We evaluate the proposed PedestrianDiffusion (PD) framework through comprehensive experiments across multiple public datasets. To ensure fair comparisons, we adopt the standard evaluation protocols established by \cite{6385773} and widely followed in recent literature~\cite{9196860,liu2020tlio,teng2025diffusionimu}, with specific metrics detailed in \autoref{subsec:analysis_metrics}. Furthermore, to conserve computational resources while maintaining consistent assessment across diverse baselines, the results reported in \autoref{tab:cross_method_comparison} and \autoref{tab:sota_comparison} are derived from models trained exclusively on the RoNIN dataset~\cite{9196860}.

\subsection{Analysis Metrics}
\label{subsec:analysis_metrics}
\noindent \textbf{Absolute Trajectory Error}
\input{equation/ATE}
Proposed by \cite{6385773}, the Absolute Trajectory Error (ATE, \autoref{eq:ate}) quantifies global drift by calculating the Euclidean discrepancy between the estimated and ground-truth positions at the terminal time step.

\noindent \textbf{Relative Trajectory Error}
\input{equation/RTE}
The Relative Trajectory Error (RTE, \autoref{eq:rte}) \cite{6385773} evaluates local consistency by averaging positional errors over fixed distance intervals, isolating systemic drift from the total trajectory length.

\noindent \textbf{Total Length Ratio}
\input{equation/TLR}
The Total Length Ratio (TLR, \autoref{eq:tlr}) measures path scale consistency. Deviations from unity characterize the error modality: values $>1.0$ imply high-frequency jitter, whereas values $<1.0$ indicate systematic motion underestimation (oversmoothing).

\noindent \textbf{Mean Cosine Similarity}
\input{equation/MCS}
The Mean Cosine Similarity (MCS, \autoref{eq:mcs}) assesses directional consistency. An MCS of $1.0$ implies perfect angular alignment, while lower values reflect trajectory drift, angular deviation, or high-frequency orientation jitter.

\subsection{Performance Analysis}

\noindent \textbf{Comparison with Baseline Approaches.}
\label{sec:cross_method_comparison}
\input{table/cross_method_comparison}
\autoref{tab:cross_method_comparison} reports evaluations on the RoNIN dataset~\cite{9196860} solely. For parity, we replicated the architectures of IoNet~\cite{Chen_Lu_Markham_Trigoni_2018}, RoNIN (ResNet, TCN, LSTM)~\cite{9196860}, TLIO~\cite{liu2020tlio}, and LLIO~\cite{wang2022llio}, extending all outputs to 3D space. Traditional recurrent and filter-based baselines exhibit substantial trajectory drift, yielding high ATE and RTE across users. While training exclusively on a single domain (PD Single Domain) artificially suppresses absolute positional errors (ATE of \qty{2.23}{m} seen, \qty{2.25}{m} unseen), it severely degrades trajectory shape retention and kinematic realism, yielding poor TLR ($\sim$\num{0.65}) and MCS ($<$\num{0.15}).

Conversely, our full diffusion formulation, \textbf{PD (Ours)}, balances positional accuracy with kinematic fidelity. In known users (RoNIN Seen), it achieves the lowest ATE (\qty{3.17}{m}) and RTE (\qty{2.24}{m}) among generalized models, alongside a near-perfect TLR of \num{0.99}. PD also demonstrates robust zero-shot generalization (RoNIN Unseen), securing the best overall ATE (\qty{4.82}{m}) while maintaining competitive RTE (\qty{3.71}{m}) and TLR (\num{0.93}) metrics. Across all scenarios, PD maintains stable MCS values (\num{0.75}--\num{0.80}), confirming that beyond minimizing absolute numerical error, our spectral formulation synthesizes kinematically realistic and temporally coherent trajectories.

\noindent \textbf{Diffusion versus Deterministic Regression.}
To isolate the contribution of the generative formulation, \autoref{tab:cross_method_comparison} compares our proposed single-step diffusion model against deterministic regression baselines trained with identical loss objectives. Replacing the spectral diffusion process with a direct regression mapping (`PD Regression') systematically degrades the ATE from \qty{3.17}{m} to \qty{3.74}{m} on seen users, and from \qty{4.82}{m} to \qty{5.69}{m} in zero-shot unseen users. Furthermore, our full dual-domain generative model (`PD (Ours)') consistently outperforms both its time-domain diffusion counterpart (`PD Time') and time-domain regression (`PD Time Regr.'). These results confirm that the diffusion-based training objective, combined with spectral preconditioning, imparts superior structural priors for robust state estimation compared to standard sequence-to-sequence regression.

\noindent \textbf{Comparison with State-of-the-Art.}
\input{table/comparison_SOTA}
\autoref{tab:sota_comparison} benchmarks PD against specialized 2D State-of-the-Art (SOTA) methods on the RoNIN dataset~\cite{9196860}, utilizing strictly IMU observations. To ensure a fair evaluation against the planar models EqNIO~\cite{jayanth2025eqnio} and DiffusionIMU~\cite{teng2025diffusionimu}, we project our 6D predictions onto the 2D horizontal plane. Under this unified protocol, PD achieves the lowest positional errors on seen users (ATE \qty{3.15}{m}, RTE \qty{2.23}{m}). On unseen users, while EqNIO demonstrates strong generalization in absolute trajectory (ATE \qty{4.26}{m}), PD achieves a superior relative trajectory error (RTE \qty{3.70}{m} versus EqNIO's \qty{4.57}{m}) and substantially outperforms DiffusionIMU (improving unseen ATE from \qty{5.27}{m} to \qty{4.79}{m}). Crucially, whereas EqNIO and DiffusionIMU are inherently restricted to 2D velocity regression ($\vec{\mathbf{v}}$), PD performs holistic 6D state estimation ($\vec{\mathbf{v}}, \boldsymbol{\vec{\omega}}$). Training this comprehensive 6D model requires \qty{10}{h} on two consumer-grade RTX 3090 GPUs; although this trails EqNIO's lightweight \qty{1.4}{h} cycle, it is highly compute-efficient compared to the \qty{10}{h} required by DiffusionIMU on an enterprise-grade NVIDIA A100. Finally, identical pre-trained weights are utilized across both \autoref{tab:cross_method_comparison} and \autoref{tab:sota_comparison}, with metric variations stemming entirely from the 2D spatial projection.

\noindent \textbf{Comparison Across Datasets.}
\input{table/performance_datasets}
\autoref{tab:PedestrianDiffusion_performance} details the evaluation across the OxIOD~\cite{chen2018oxioddatasetdeepinertial}, RIDI~\cite{yan2017ridirobustimudouble}, RoNIN~\cite{9196860}, and TLIO~\cite{liu2020tlio} datasets under a single-weight joint training regime, denoted as the hybrid dataset setting. This protocol highlights the performance nuances introduced by diverse ground-truth (GT) acquisition systems. PD demonstrates robust generalization on TLIO (Orientation ATE \qty{0.12}{\radian}), which utilizes a rigorous visual-inertial setup for precise 3D tracking. Performance variations across other benchmarks directly reflect the fidelity of their supervision signals; for instance, the Position RTE disparity within OxIOD (VICON \qty{0.30}{m} versus Tango \qty{1.34}{m}) underscores the precision gap between optical motion capture and SLAM-based annotations. Notably, the TLR remains proximal to unity across all domains, demonstrating that PD consistently preserves accurate trajectory scale invariant to the underlying GT modality.

\noindent \textbf{Inference Benchmark.}
\begin{figure*}
    \centering 
    \includegraphics[width=0.245\linewidth]{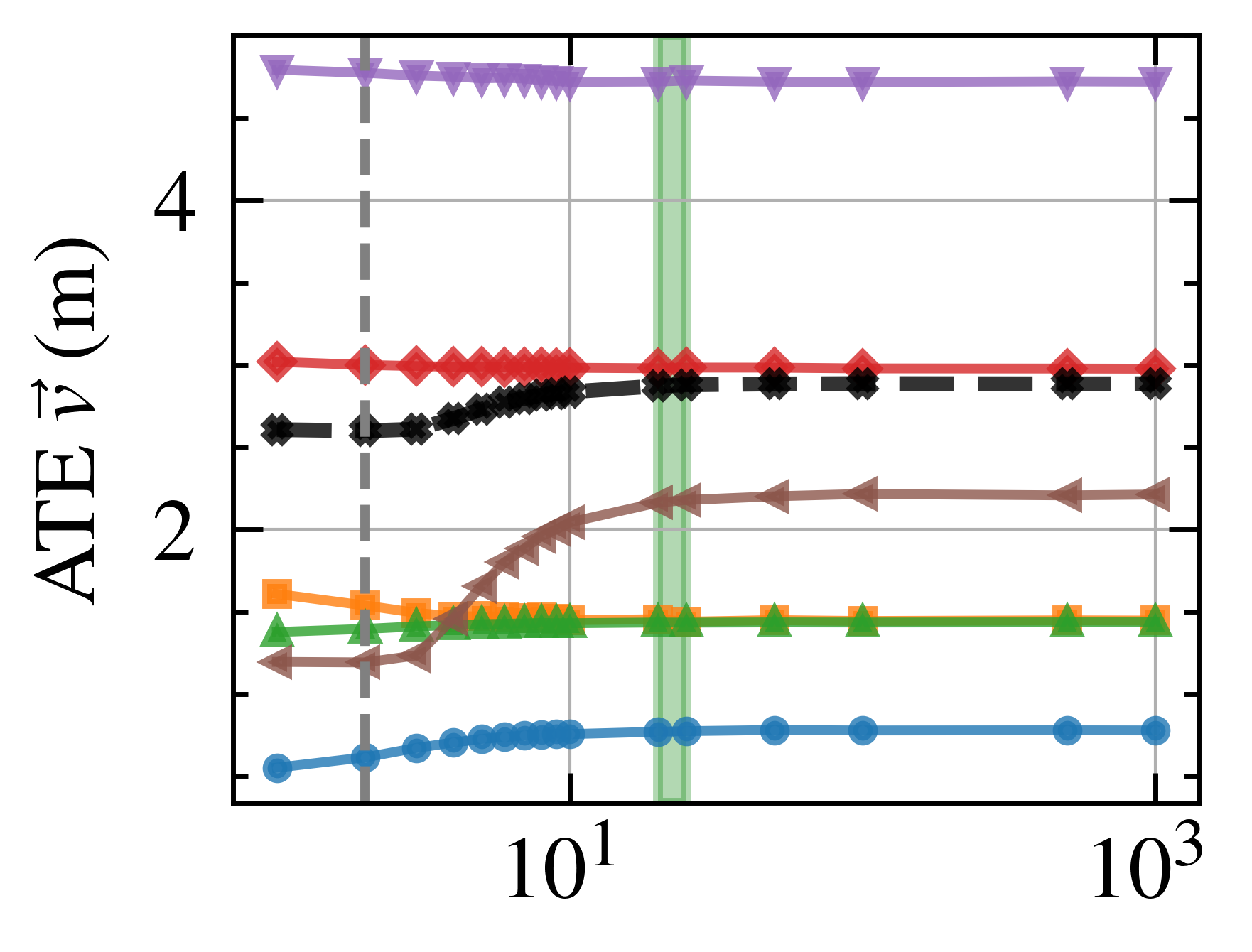}
    \includegraphics[width=0.245\linewidth]{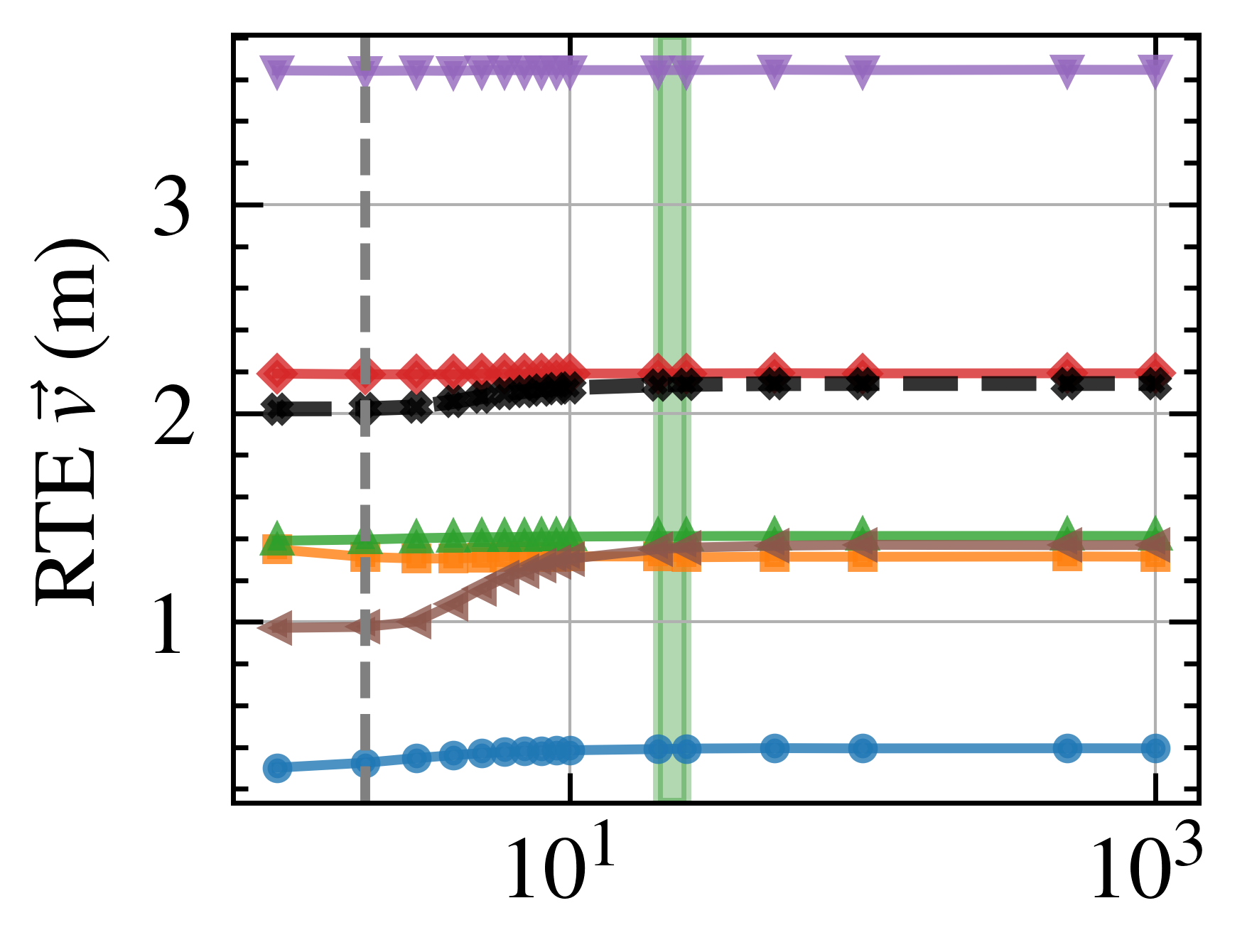}
    \includegraphics[width=0.245\linewidth]{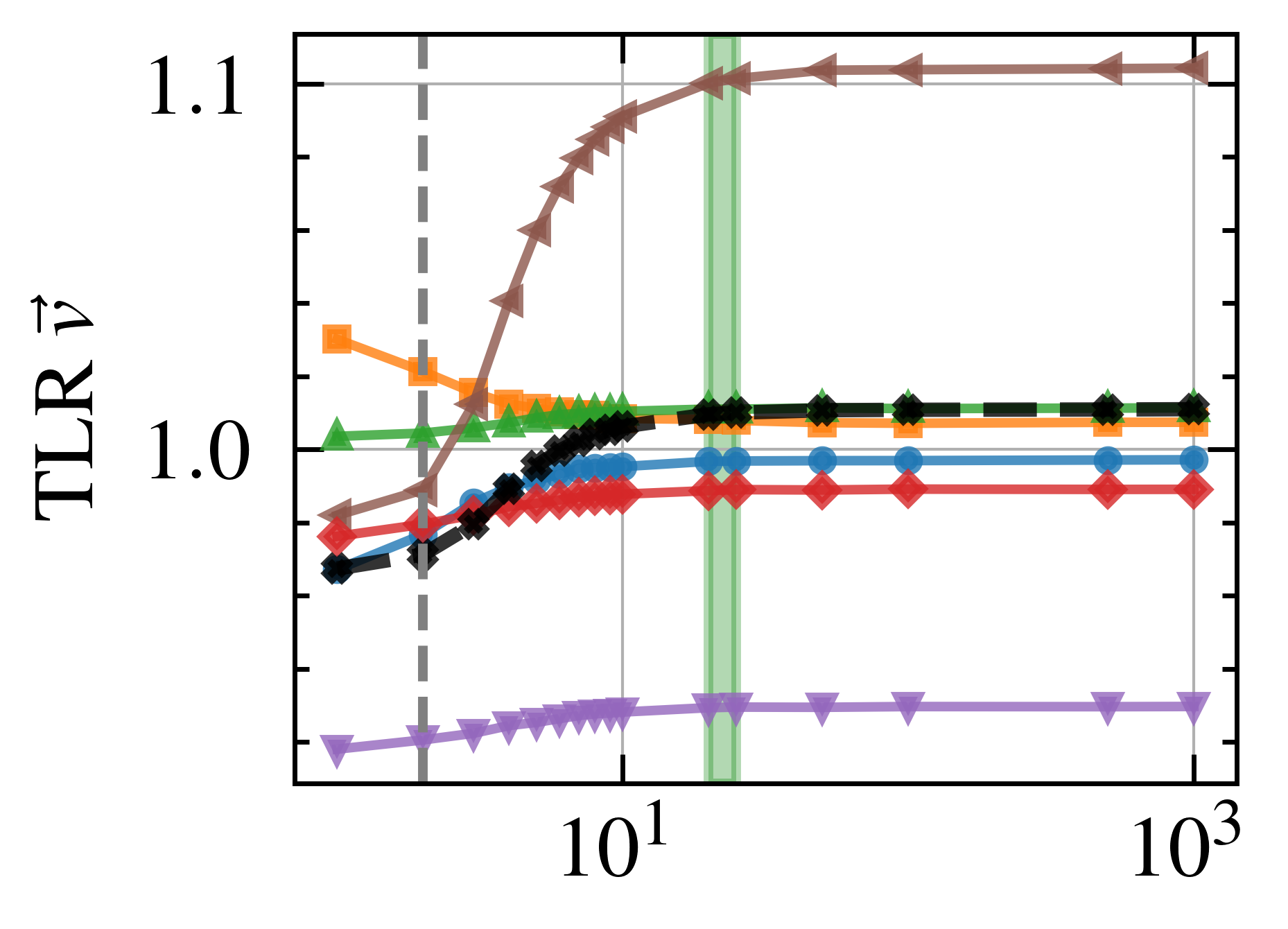}
    \includegraphics[width=0.245\linewidth]{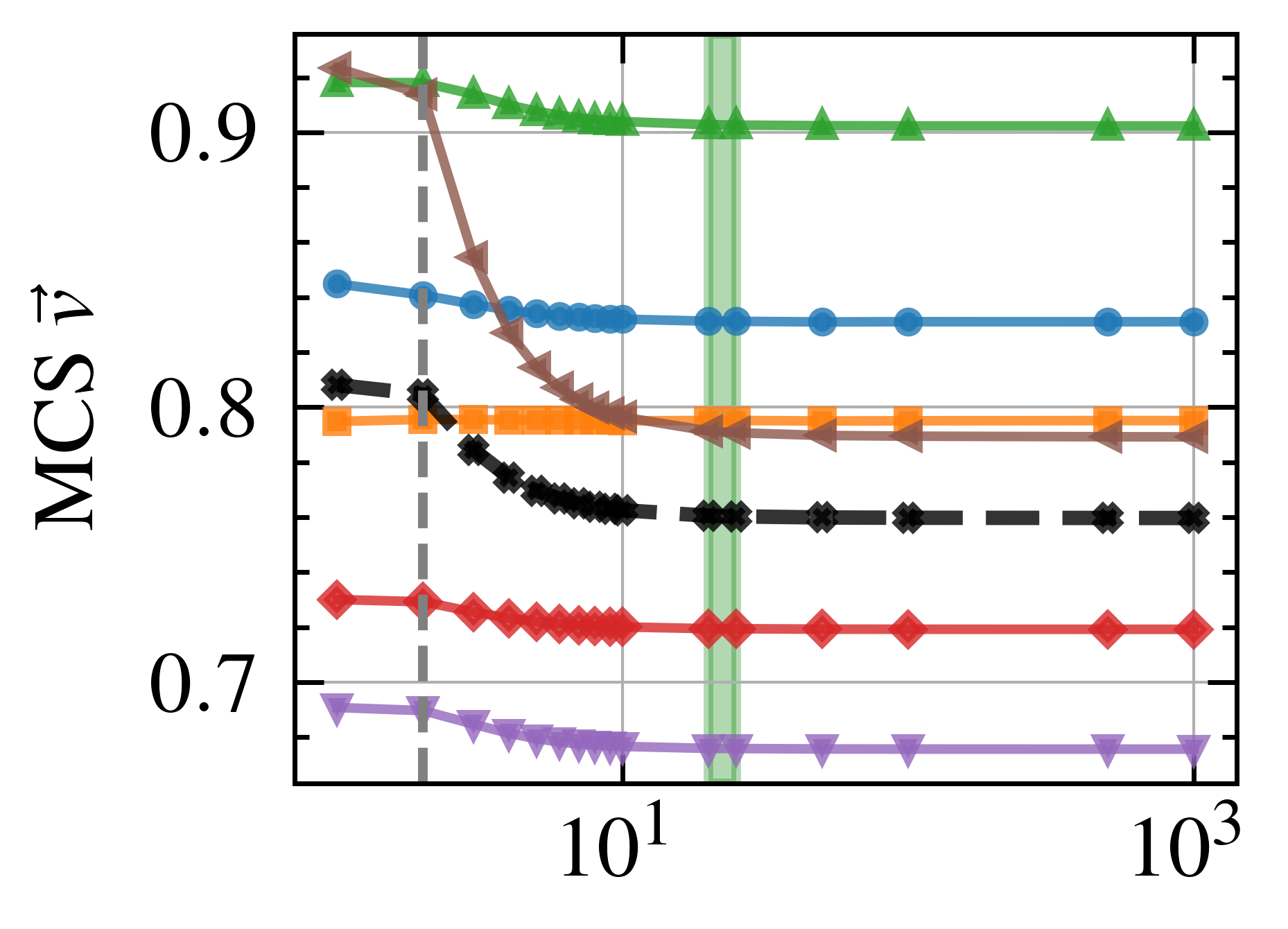}

    \includegraphics[width=0.245\linewidth]{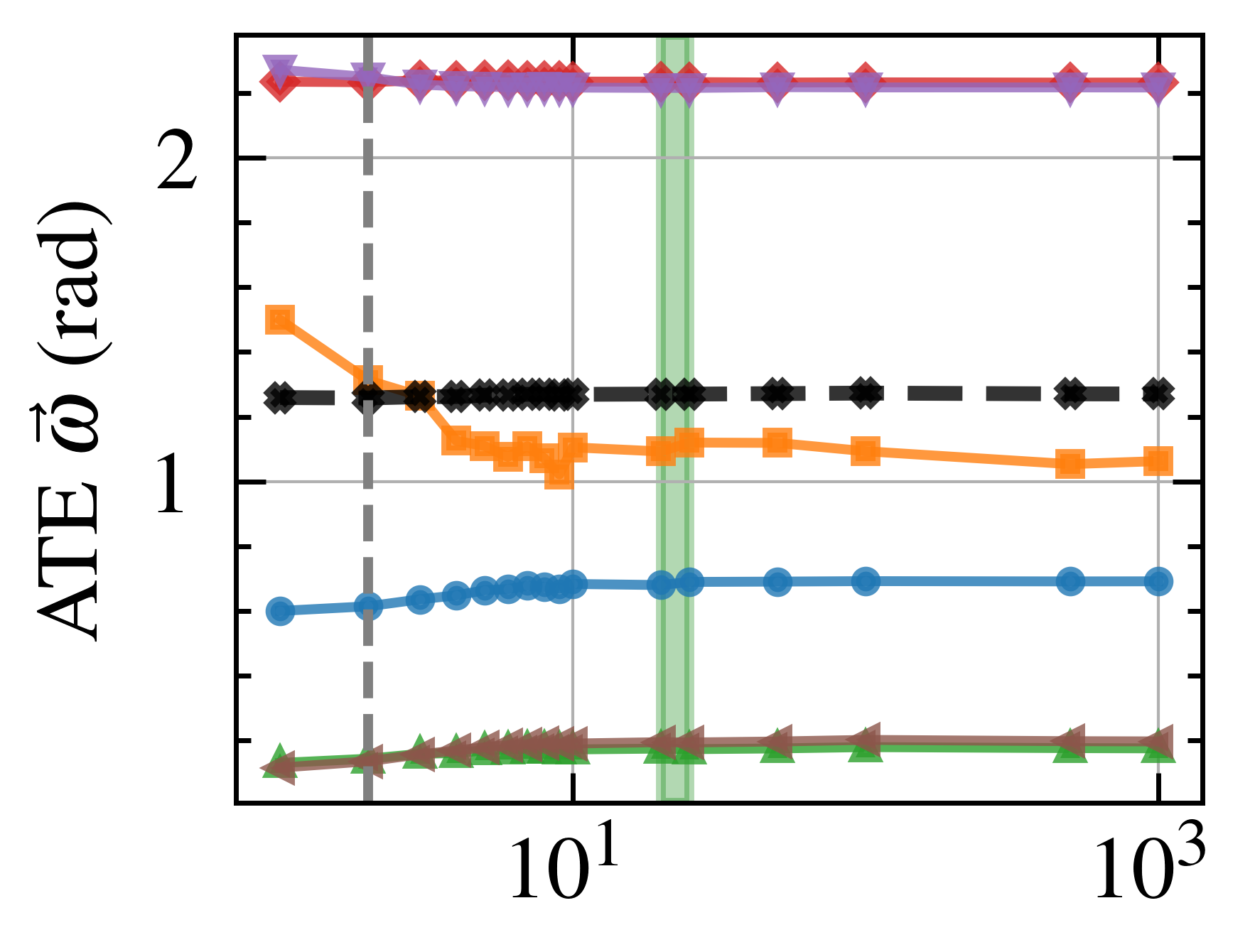}
    \includegraphics[width=0.245\linewidth]{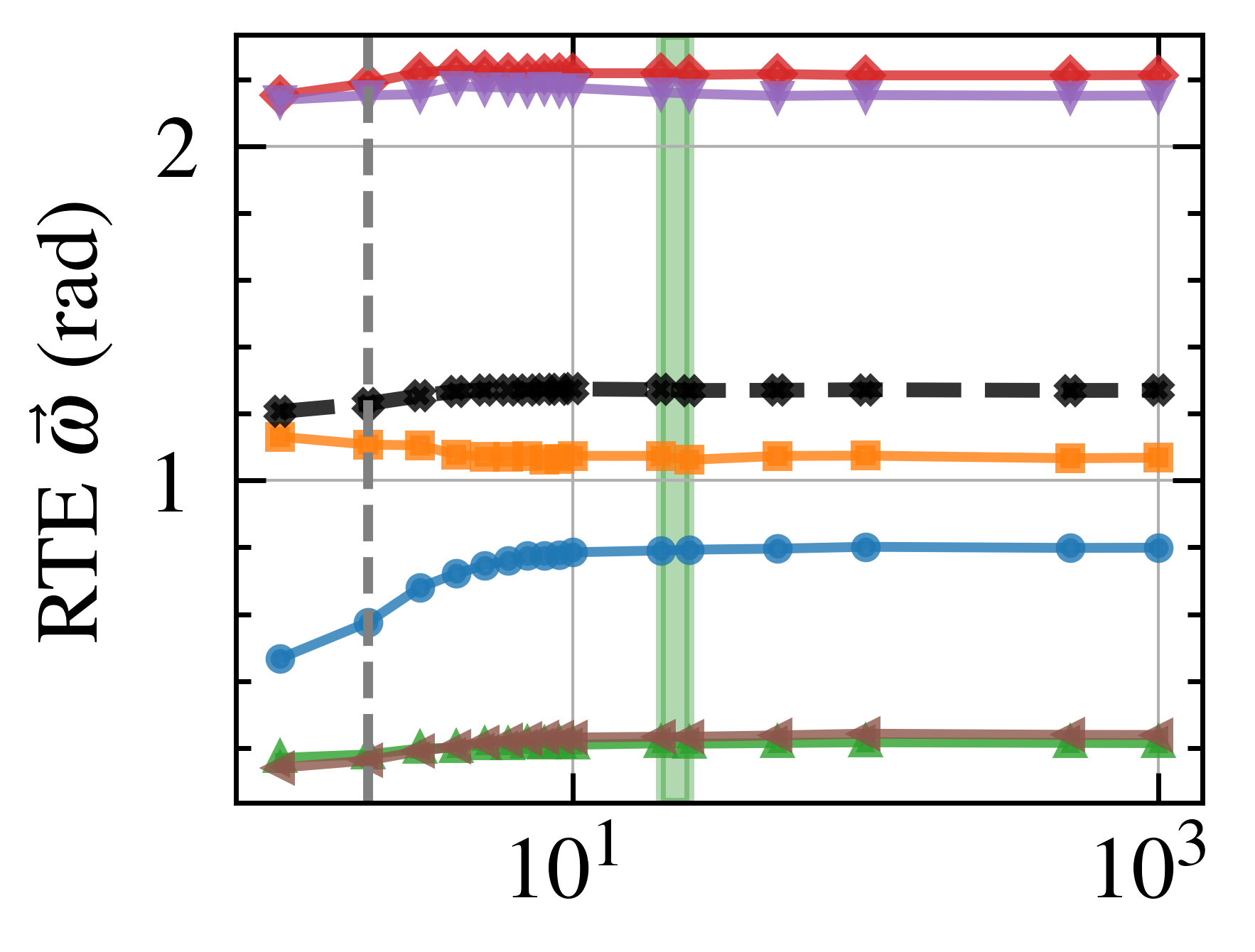}
    \includegraphics[width=0.245\linewidth]{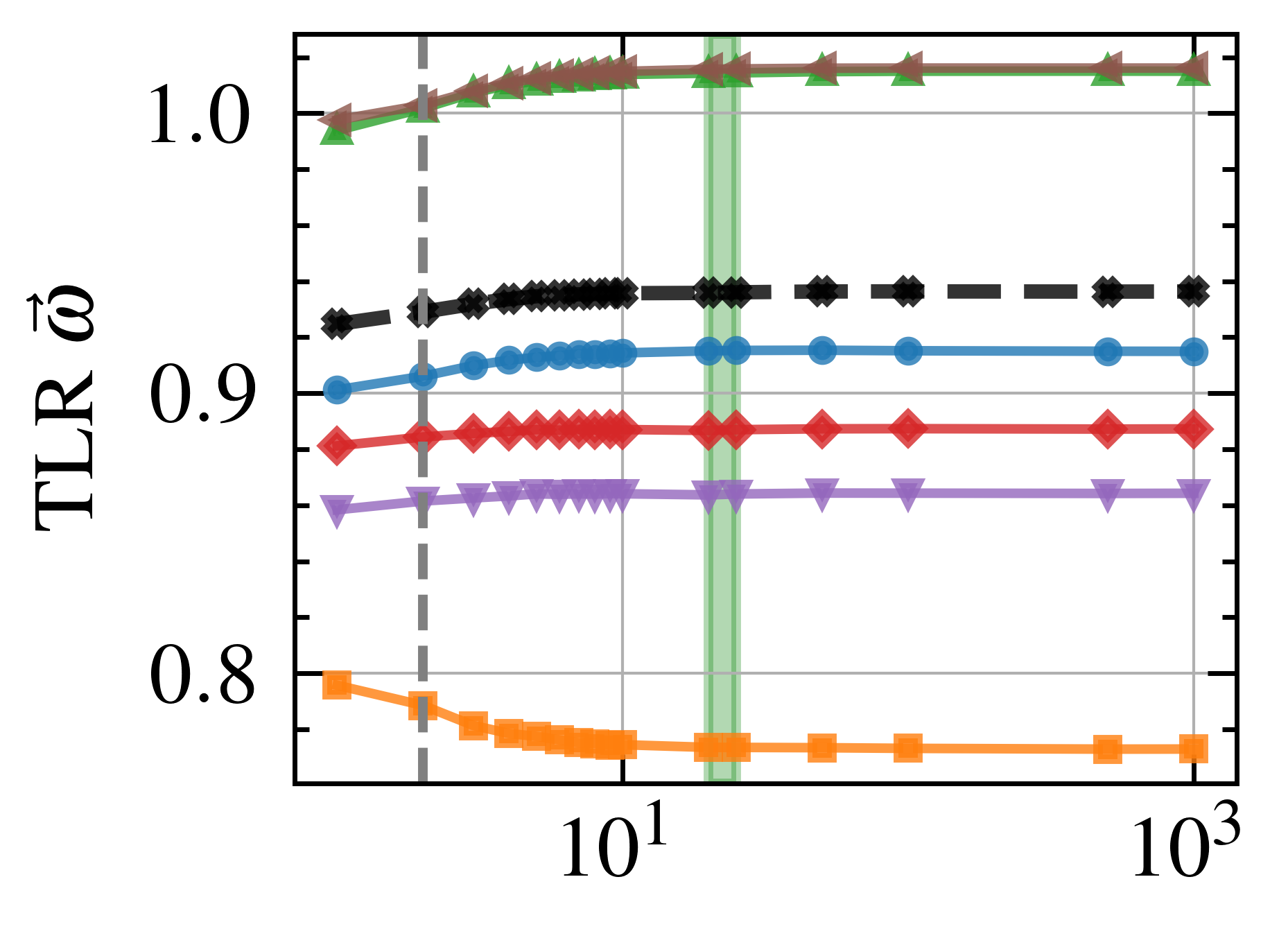}
    \includegraphics[width=0.245\linewidth]{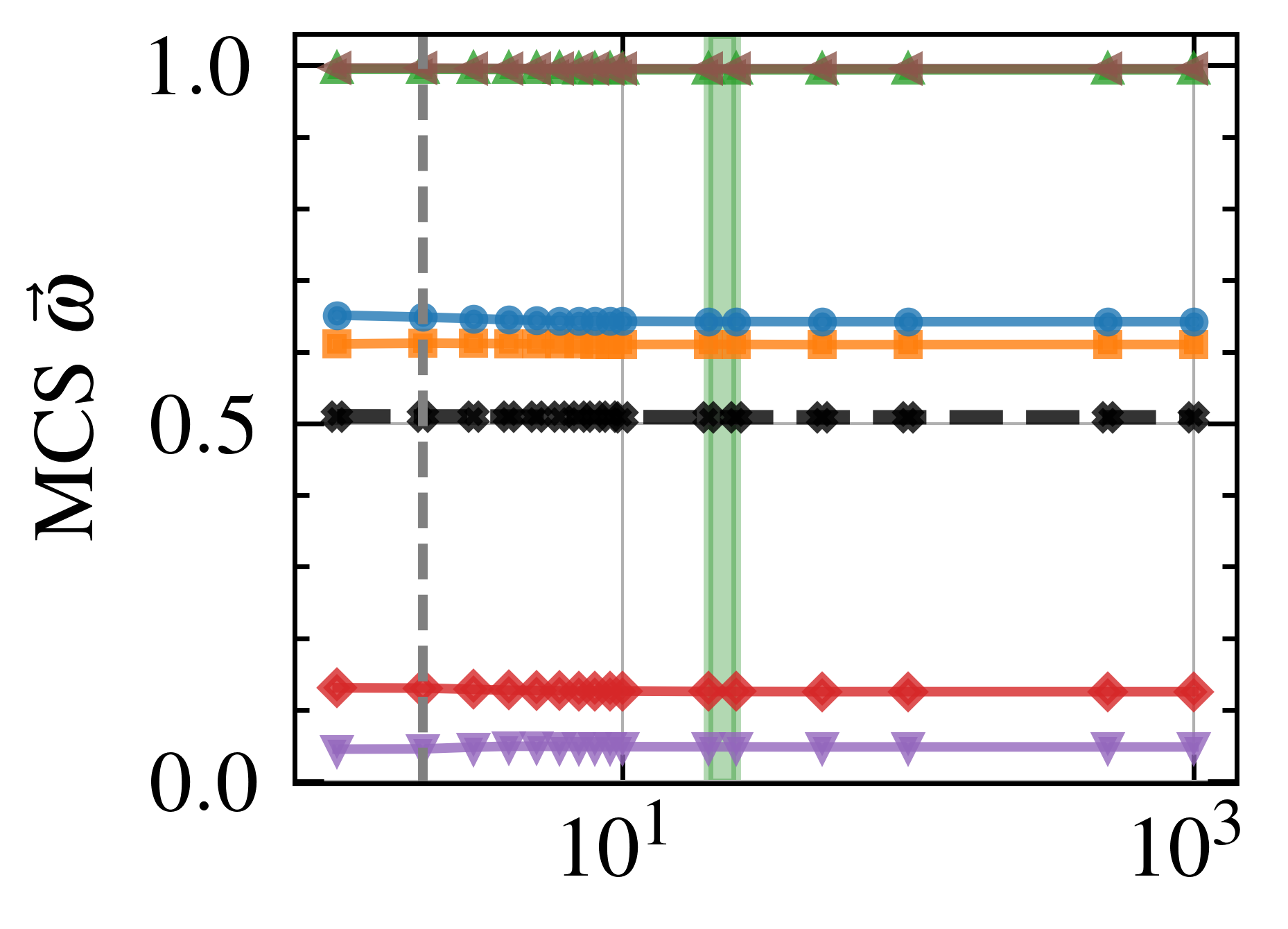}

    \includegraphics[width=\linewidth]{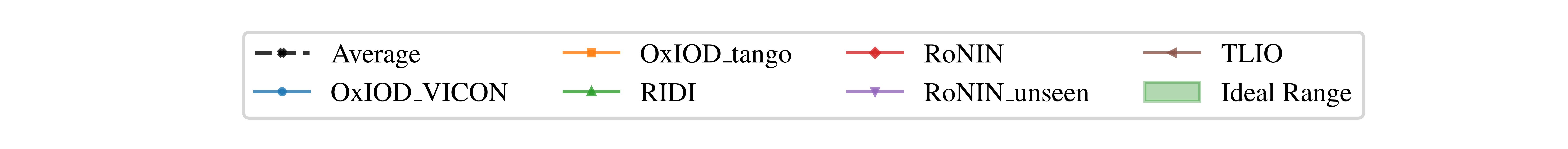}
    \caption{The figures illustrate the inference steps VS. metrics performance across datasets. The green area is the optimal range for the DPM-Solver++ scheduler 20--25 steps.}
    \label{fig:steps_inference}
\end{figure*}

We evaluate the sensitivity of trajectory accuracy to the number of diffusion inference steps, $T$. As demonstrated in \autoref{fig:steps_inference}, both ATE and RTE degrade as $T$ increases, a behavior that diverges from the standard $20$--$25$ step convergence typically required in image generation~\cite{lu2025dpm}. We attribute this degradation to \textit{diffusion hallucination}: excessive iterative refinement over-optimizes predictions toward the learned prior, introducing physically ungrounded kinematic artifacts. Notably, tracking errors remain sufficiently low at $T=1$. This demonstrates that a single inference step can optimally recover the trajectory manifold and suppress hallucination-induced drift, concurrently reducing the computational overhead.

\noindent \textbf{Computational Analysis.}
\input{table/computation}
\autoref{tab:computation} evaluates the computational complexity and deployment efficiency of all methods utilizing an NVIDIA RTX 3090 GPU and a Raspberry Pi 5 edge CPU. Unlike discriminative baselines that compress a temporal window into a single motion estimate ($N=1$), PD performs dense state estimation to synthesize a physically consistent 100-frame trajectory in parallel ($N=100$). This necessitates a higher capacity architecture (\qty{37.26}{M} parameters) and incurs a higher absolute computational cost (\qty{23.37}{GFLOPs}).

Despite this absolute complexity, PD demonstrates highly favorable amortized efficiency. On the RTX 3090, PD generates a complete 100-frame sequence in \qty{107.71}{ms}, yielding an amortized latency of \qty{1.08}{ms} per frame. While trailing the lightweight RoNIN-TCN~\cite{9196860} and RoNIN-LSTM~\cite{9196860} (\qty{0.04}{ms/frame}), it significantly outperforms single-output regression models: IoNet (\qty{3.54}{ms/frame}), LLIO (\qty{5.85}{ms/frame}), and TLIO~\cite{liu2020tlio} (\qty{6.93}{ms/frame}). On the Raspberry Pi 5, PD achieves an amortized edge latency of \qty{9.88}{ms/frame} while maintaining a lower peak memory footprint (\qty{1.83}{GB}) than all baselines (\qty{2.37}{GB}--\qty{2.45}{GB}). Given the substantial gains in localization accuracy and kinematic fidelity (mitigating the severe scale distortion and poor TLR of lightweight models), this computational trade-off is justified, proving that generative models are viable for resource-constrained edge platforms.

\begin{figure}
    \centering
    \includegraphics[width=3.5in]{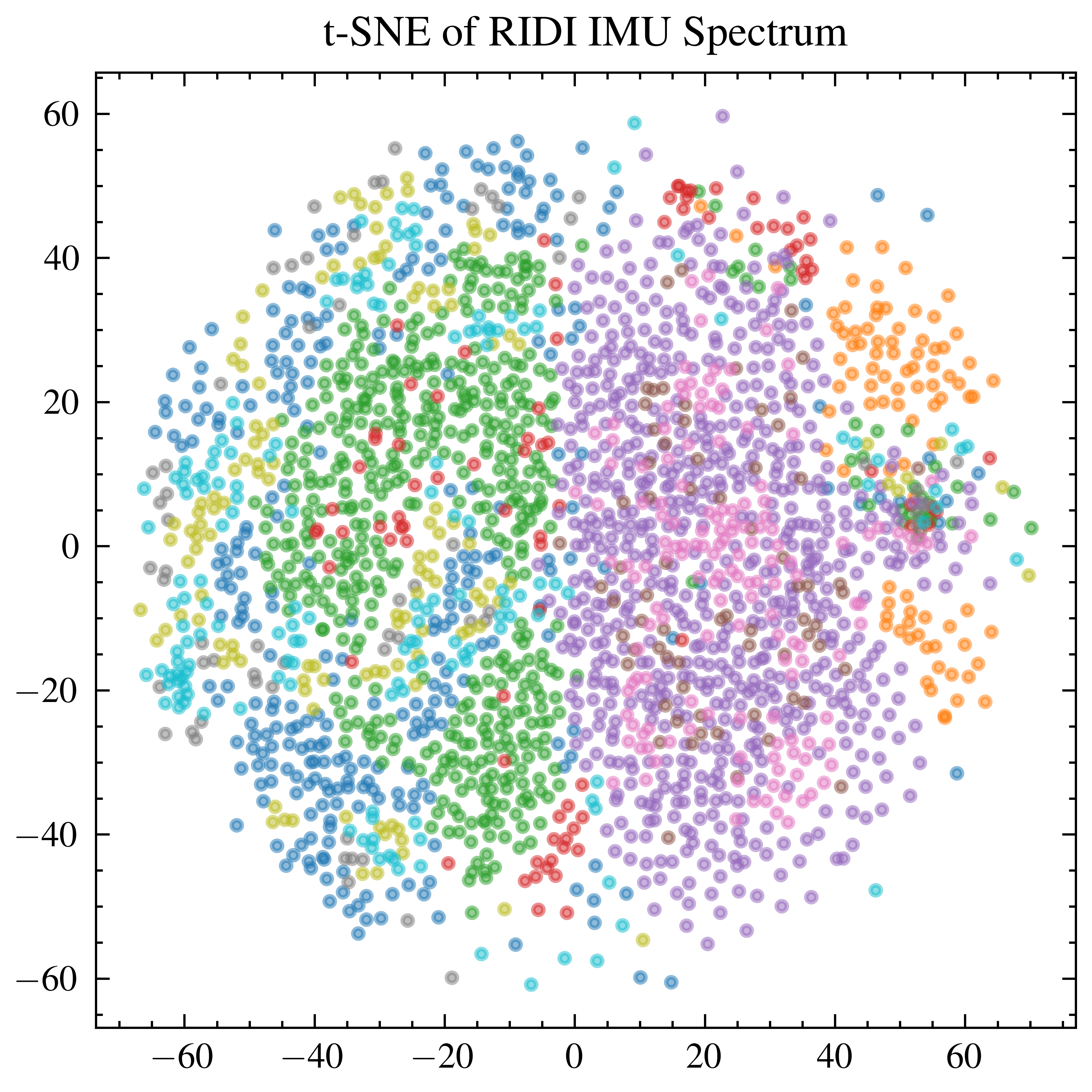}
    \includegraphics[width=3.5in]{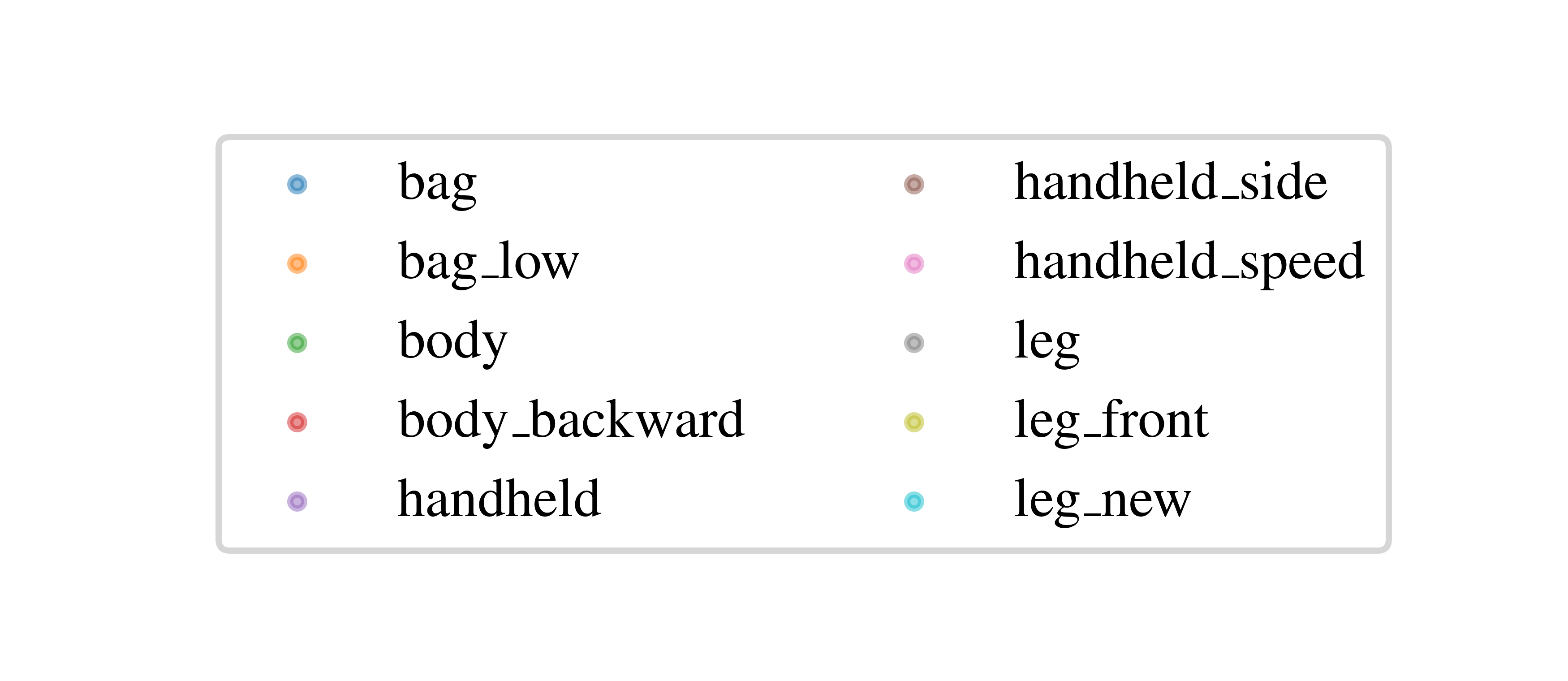}
    \caption{t-SNE visualization of raw spectrum for RIDI user profiles. The plot demonstrates that semantically similar profiles (e.g., ``handheld'' and ``handheld speed'') cluster closely in the embedding space.}
    \label{fig:tsne_clip}
\end{figure}

\noindent \textbf{CLIP Conditioning Ablation.}
\input{table/ablation_clip_conditioning}
As detailed in \autoref{tab:ablation_clip_conditioning}, we perform CLIP text encoding, with known cases and unknown cases on the RIDI~\cite{yan2017ridirobustimudouble} benchmark. In the unknown case, the keywords are placed with `unknown' as shown in \autoref{fig:sentence}. The report shows reducing orientation and positional ATE by \qty{54}{\%} ($\num{0.18} \rightarrow \qty{0.08}{rad}$) and \qty{23}{\%} ($\num{1.70} \rightarrow \qty{1.31}{m}$), respectively. This improvement validates the utility of CLIP's shared semantic space for zero-shot generalization: by embedding unobserved noise profiles in close proximity to physically similar, known configurations (e.g., mapping a novel device to a known ``handheld'' profile), the model robustly adapts to unseen motion dynamics without explicit retraining. \autoref{fig:tsne_clip} visualizes the t-SNE embedding of raw spectral features for RIDI user profiles, demonstrating that semantically similar profiles cluster closely in the embedding space, thereby facilitating effective generalization through semantic proximity. As shown in \autoref{fig:tsne_clip}, some of the different mounting positions (e.g., ``handheld'' and ``handheld speed'') are partially clustered closely in the embedding space, which necessitates CLIP to leverage the semantic similarity to further distinguish them.

\section{Discussion \& Limitations}
\noindent \textbf{Application Realistic Latency versus Hard Real-Time Execution.}
Existing approaches mentioned in \autoref{sec:related_work} rely on posterior processing, incurring a prohibitive latency of \qty{1}{\second}. Although our model achieves an amortized processing time of \qty{9.88}{ms} per frame (processing a 100-frame window in \qty{987.79}{ms}), a critical distinction must be made between offline batch throughput and hard real-time execution. In closed-loop navigation, causal dependencies dictate that state estimation cannot incorporate future frames. Given our sliding window hop size of \qty{1}{\second} (100 samples), strict sequential execution on a standard edge CPU (e.g., Raspberry Pi 5) would exceed this temporal constraint, potentially inducing buffer overflow. Therefore, while our $T=1$ convergence drastically mitigates the computational overhead inherent to multi-step diffusion, deploying PedestrianDiffusion for strictly real-time continuous navigation currently mandates discrete hardware accelerators. Ultimately, the reported \qty{9.88}{ms} amortized latency validates the architecture's theoretical throughput capacity, establishing a pathway toward---rather than an immediate realization of---CPU-bound low-latency edge operability. The issue also applies to all baselines and SOTA methods, given that all approaches are posterior-processing methods. Hence, the latency within next observation window (e.g., \qty{1}{\second}) is the same for all methods.

\noindent \textbf{Limits of Cross-Domain Generalization.}
Despite substantial improvements in trajectory consistency, residual variance in orientation estimates delineates the boundaries of static semantic conditioning. While multimodal embeddings successfully mitigate global device heterogeneity, they cannot entirely compensate for disjoint, high-frequency sensor noise profiles. This suggests that the remaining orientation errors propagate from low-level stochastic distributional shifts rather than high-level semantic discrepancies. Consequently, achieving robust zero-shot orientation generalization likely necessitates complementary paradigms, such as test-time adaptation, to dynamically align sensor distributions during inference.

\noindent \textbf{Computational Capacity versus Localization Precision.}
PedestrianDiffusion navigates the fundamental trade-off between architectural capacity and state estimation precision. Although the \qty{37.26}{M} parameter backbone is substantially heavier than discriminative regression baselines, this representational density is strictly necessary to model the full spectral bandwidth of pedestrian kinematics without underfitting complex motion manifolds. Crucially, our amortized efficiency benchmarks (\qty{1.08}{ms/frame} on GPU) demonstrate that this architectural scale does not preclude high-throughput operation, effectively decoupling model capacity from inference latency.

\noindent \textbf{Contextual Semantic Conditioning.}
Our current implementation relies on uniform prompt templates, which may not optimally represent individualized kinematic contexts. Future investigations will focus on fine-grained conditioning strategies, such as dynamically generating user-specific contextual descriptions. By more thoroughly exploiting the latent semantic space of the CLIP encoder, this approach aims to capture nuanced variations in stochastic sensor noise and motion dynamics across disparate operators and mounting configurations, thereby further enhancing zero-shot generalization.

\section{Conclusion}
We introduced PedestrianDiffusion, a novel spectral-domain generative framework for inertial navigation. By reformulating trajectory estimation as a conditional denoising process, our approach successfully recovers the high-frequency kinematic details systematically suppressed by traditional MSE-based regression paradigms, achieving state-of-the-art performance across multiple standard benchmarks. Furthermore, our empirical analyses challenge the prevailing consensus that generative models are computationally prohibitive for edge deployment. Demonstrating an amortized per-frame latency of \qty{9.88}{ms}, our framework establishes a high-precision, viable alternative to lightweight sequential filters.

Future research trajectories include:
\begin{itemize}
    \setlength\itemsep{0em}
    \setlength\parskip{0em}
    \setlength\parsep{0em}
    \item \textbf{Spectral Hardware Acceleration:} Co-designing hardware architectures optimized for STFT processing to minimize the energy footprint of spectral backbones.
    \item \textbf{Adaptive Inference Scheduling:} Dynamically modulating the number of diffusion steps ($T$) contingent on instantaneous kinematic complexity to optimize the energy-accuracy Pareto front.
    \item \textbf{Universal Motion Foundation Models:} Extending multimodal semantic conditioning to vehicular and aerial dynamics, facilitating cross-domain zero-shot generalization within a unified navigational framework.
\end{itemize}
\section{Acknowledgments}
This research was funded by the National Fire Agency (NFA) and the Korea Planning \& Evaluation Institute of Industrial Technology (KEIT) of the Republic of Korea (No. RS-2025-02313957).

The authors used Gemini (Google) to assist with language editing and improving the readability of portions of this manuscript. All technical content, analyses, interpretations, and conclusions were developed and verified by the authors, who take full responsibility for the contents of this paper.
\bibliographystyle{IEEEtran}
\bibliography{citation}
\clearpage
\input{Appendices}



\end{document}

%% file: figure/Diffusion_overview.tex
\begin{figure}
    \centering
    \includegraphics[width=\columnwidth]{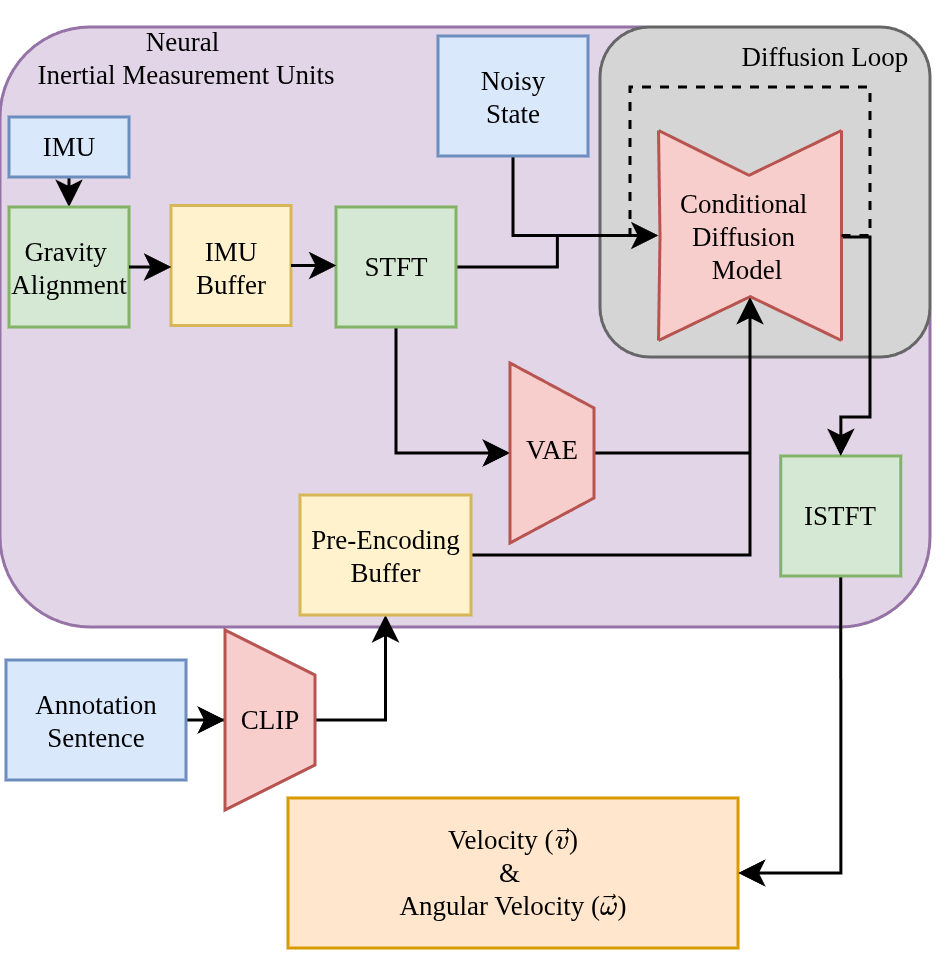}
    \caption{The overall process of the PedestrianDiffusion model. }
    \label{fig:Diffusion_overview}
\end{figure}

%% file: figure/sentence.tex
\begin{figure}
    \centering
    \includegraphics[width=\columnwidth]{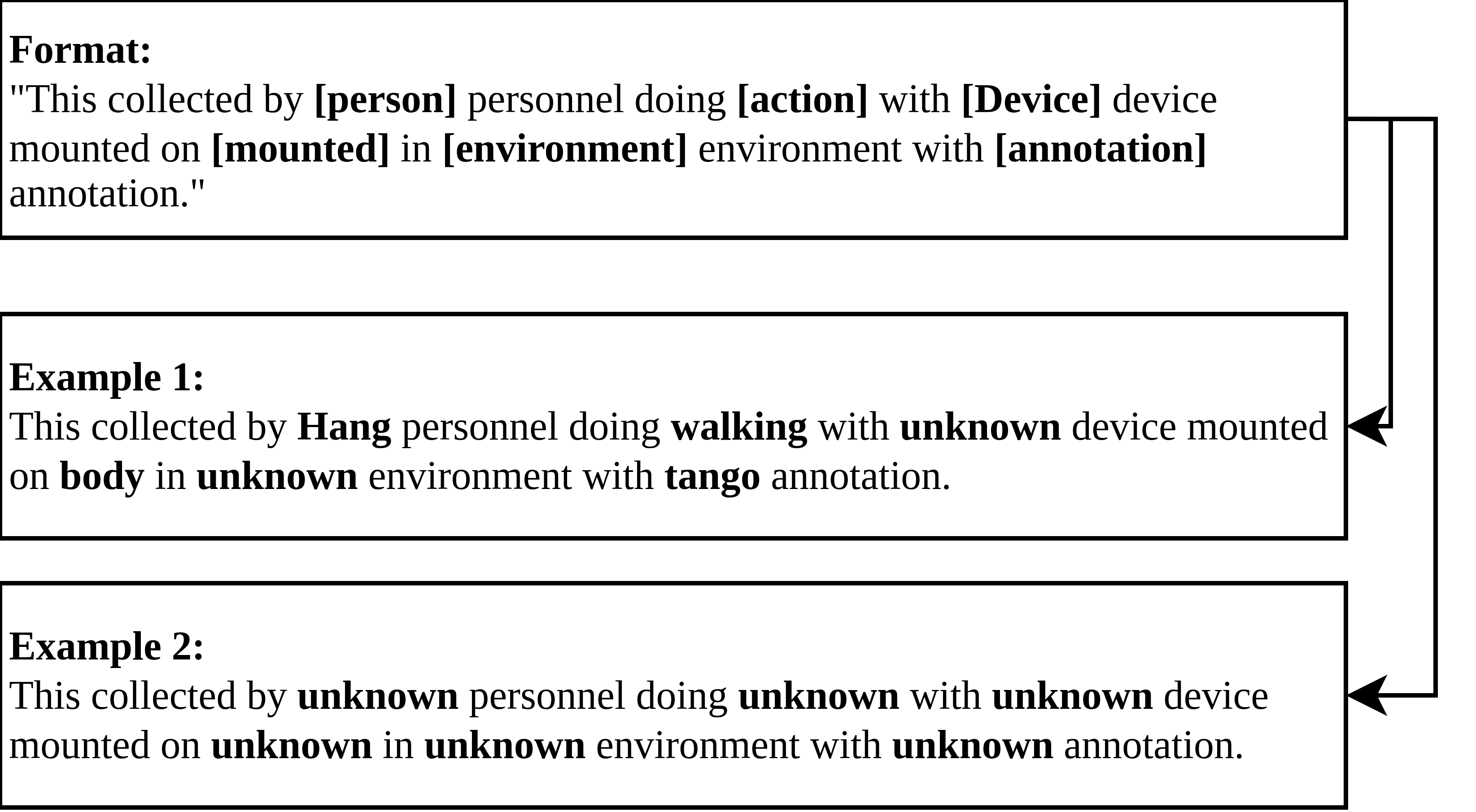}
    \caption{Natural language prompt templates used for CLIP encoding. We serialize metadata into structured sentences to generate semantic embeddings for each dataset. ``Example 1'' illustrates a known configuration from the RIDI dataset, while ``Example 2'' demonstrates the template for unknown or unannotated scenarios. We iterate through all attribute combinations to prevent overfitting, using an ``unknown'' token as a fallback for unseen domains during inference.}
    \label{fig:sentence}
\end{figure}

%% file: equation/loss_function.tex
\paragraph{Total Loss with Uncertainty Weighting.}
To adaptively balance the competing objectives during training, we utilize homoscedastic uncertainty weighting. The total loss function, denoted as $\mathcal{L}_{\text{total}}$, aggregates individual loss terms $\ell_{m,k}$ across two sensor modalities and five distinct tasks. Specifically, let $m \in \{\vec{\mathbf{v}}, \boldsymbol{\vec{\omega}}\}$ represent the modality (where $\vec{\mathbf{v}}$ denotes velocity and $\boldsymbol{\vec{\omega}}$ denotes angular velocity), and let $k \in \{1, \dots, 5\}$ index the specific objective component (e.g., time-domain reconstruction, frequency alignment). We introduce a learnable log-variance parameter $s_{m,k}$ for each task-modality pair to regulate the trade-off between loss magnitude and model uncertainty. The final objective is defined as:

\begin{equation}
    \mathcal{L}_{\text{total}} = \sum_{m \in \{\vec{\mathbf{v}}, \boldsymbol{\vec{\omega}}\}} \sum_{k=1}^{5} \left( \exp(-s_{m,k}) \cdot \ell_{m,k} + s_{m,k} \right)
\end{equation}

\noindent where the term $\exp(-s_{m,k})$ acts as an adaptive precision weight for the raw loss $\ell_{m,k}$, and the additive term $s_{m,k}$ serves as a regularizer to prevent the model from predicting infinite uncertainty.

\paragraph{Robust Reconstruction Objective.}
The component losses utilize the Huber loss function $H_{\delta}(x, y)$ with threshold $\delta$ as standard Huber loss:

\begin{equation}
    H_{\delta}(x, y) =
    \begin{cases}
        \frac{1}{2}(x - y)^2                 & \text{if } |x - y| \le \delta \\
        \delta (|x - y| - \frac{1}{2}\delta) & \text{otherwise}
    \end{cases}
\end{equation}

\paragraph{Component Objectives.}
Let $\hat{\mathbf{y}}_t, \mathbf{y}_t$ denote the time-domain predictions and targets, and $\hat{\mathbf{y}}_f, \mathbf{y}_f$ denote their frequency-domain counterparts (via STFT). The individual loss components are defined as follows:
\begin{align}
    \label{eq:freq_recon}
    \ell_{m,1} & = H_\delta\Big(   \hat{\mathbf{y}}_f, \,   \mathbf{y}_f \Big)                               \\
    \label{eq:integral_consist}
    \ell_{m,2} & = H_\delta\bigg(   \sum \hat{\mathbf{y}}_t \Delta t, \,   \sum \mathbf{y}_t \Delta t \bigg) \\
    \label{eq:time_recon}
    \ell_{m,3} & = H_\delta\Big(   \hat{\mathbf{y}}_t, \,   \mathbf{y}_t \Big)                               \\
    \label{eq:time_sim}
    \ell_{m,4} & = H_\delta\Big(   \mathcal{S}(\hat{\mathbf{y}}_t, \mathbf{y}_t), \, 0 \Big)                 \\
    \label{eq:freq_sim}
    \ell_{m,5} & = H_\delta\Big(   \mathcal{S}(\hat{\mathbf{y}}_f, \mathbf{y}_f), \, 0 \Big)
\end{align}

\paragraph{Explanations of Objectives:}
\begin{itemize}
    \setlength\itemsep{0em}
    \setlength\parskip{0em}
    \setlength\parsep{0em}
    \item \textbf{Frequency Reconstruction ($\ell_{m,1}$, \autoref{eq:freq_recon}):} Minimizes the spectral density error, ensuring the model captures the correct periodicity and vibration patterns of the inertial signal.
    \item \textbf{Integral Consistency ($\ell_{m,2}$, \autoref{eq:integral_consist}):} Minimizes the error of the signal integral (velocity/orientation). This imposes a physical constraint that prevents the accumulation of drift over time.
    \item \textbf{Time Reconstruction ($\ell_{m,3}$, \autoref{eq:time_recon}):} A direct reconstruction loss on the raw waveform. We apply a higher scaling factor ($10\kappa$) here to enforce strictly accurate fine-grained signal synthesis.
    \item \textbf{Time Domain Similarity ($\ell_{m,4}$, \autoref{eq:time_sim}):} Utilizes the Cosine similarity loss $\mathcal{S}$ to maximize the cosine similarity between the predicted and target embeddings in the time domain, focusing on shape and phase alignment rather than amplitude.
    \item \textbf{Frequency Domain Similarity ($\ell_{m,5}$, \autoref{eq:freq_sim}):} Analogous to $\ell_{m,4}$ but applied in the spectral domain, ensuring the distribution of dominant frequencies matches the target even if the exact phase alignment varies.
\end{itemize}

%% file: equation/ATE.tex
\begin{equation}
    \begin{aligned}
        \text{ATE} = \left( \frac{1}{N} \sum_{i=1}^{N} \| \hat{\mathbf{p}}_i - \mathbf{p}_i \|^2 \right)^{1/2}
        \label{eq:ate}
    \end{aligned}
\end{equation}
\begin{itemize}
    \setlength\itemsep{0em}
    \setlength\parskip{0em}
    \setlength\parsep{0em}
    \item \(\hat{\mathbf{p}}_i, \mathbf{p}_i \in \mathbb{R}^3\) are the estimated position (aligned to the ground truth frame) and the ground truth position at time step \(i\), respectively.
    \item \(N\) is the total number of poses in the trajectory.
    \item \(\left\| \cdot \right\|\) denotes the Euclidean norm.
\end{itemize}

%% file: equation/RTE.tex
\begin{equation}
    \begin{aligned}
        \text{RTE} = \frac{1}{N-1} \sum_{i=1}^{N-1} \lVert \Delta \hat{\mathbf{p}}_i - \Delta \mathbf{p}_i \rVert
    \end{aligned}
    \label{eq:rte}
\end{equation}

\begin{itemize}
    \setlength\itemsep{0em}
    \setlength\parskip{0em}
    \setlength\parsep{0em}
    \item \(\Delta \hat{\mathbf{p}}_i = \hat{\mathbf{p}}_{i+1} - \hat{\mathbf{p}}_i\) is the estimated displacement vector between steps \(i\) and \(i+1\).
    \item \(\Delta \mathbf{p}_i = \mathbf{p}_{i+1} - \mathbf{p}_i\) is the ground-truth displacement vector between steps \(i\) and \(i+1\).
    \item \(n\) is the total number of poses.
\end{itemize}

%% file: equation/TLR.tex
\begin{equation}
    \text{TLR} = \frac{\sum_{i=2}^{N} \| \hat{\mathbf{p}}_i - \hat{\mathbf{p}}_{i-1} \|}{\sum_{i=2}^{N} \| \mathbf{p}_i - \mathbf{p}_{i-1} \|}
    \label{eq:tlr}
\end{equation}
\begin{itemize}
    \setlength\itemsep{0em}
    \setlength\parskip{0em}
    \setlength\parsep{0em}
    \item \(\|\cdot\|\) denotes the Euclidean norm.
    \item \(N\) is the total number of frames in the sequence (resulting in \(N-1\) segments).
\end{itemize}

%% file: equation/MCS.tex
\begin{equation}
    \text{MCS} = \frac{1}{N-1} \sum_{i=2}^{N} \frac{\hat{\mathbf{d}}_i \cdot \mathbf{d}_i}{\| \hat{\mathbf{d}}_i \| \| \mathbf{d}_i \|}
    \label{eq:mcs}
\end{equation}
where $\hat{\mathbf{d}}_i = \hat{\mathbf{p}}_i - \hat{\mathbf{p}}_{i-1}$ and $\mathbf{d}_i = \mathbf{p}_i - \mathbf{p}_{i-1}$ represent the estimated and ground truth displacement vectors, respectively.
\begin{itemize}
    \setlength\itemsep{0em}
    \setlength\parskip{0em}
    \setlength\parsep{0em}
    \item \(\cdot\) denotes the dot product between two vectors.
    \item \(\|\cdot\|\) denotes the Euclidean norm (magnitude of the vector).
    \item \(N\) is the total number of frames in the sequence.
\end{itemize}

%% file: table/cross_method_comparison.tex
\begin{table*}
    \centering
    \caption{\textbf{Performance comparison on RoNIN datasets in 3D space.} Values are reported as mean $\pm$ standard deviation. \textbf{Bold} indicates the best performance. ATE and RTE (lower is better); TLR and MCS (closer to 1 is better).}
    \label{tab:cross_method_comparison}
    \small
    \setlength{\tabcolsep}{2pt}
    \begin{tabular*}{\linewidth}{l @{\extracolsep{\fill}} cccc | cccc}
        \toprule
        & \multicolumn{4}{c}{\textbf{RoNIN Seen}} & \multicolumn{4}{c}{\textbf{RoNIN Unseen}} \\
        \cmidrule(lr){2-5} \cmidrule(lr){6-9}
        \textbf{Method} & ATE (m) & RTE (m) & TLR & MCS & ATE (m) & RTE (m) & TLR & MCS \\
        \midrule
        \multicolumn{9}{l}{\textit{Baseline Comparison}} \\
        IoNet & 4.08 $\pm$ 1.89 & 2.66 $\pm$ 1.35 & 0.97 $\pm$ 0.04 & -- & 5.64 $\pm$ 4.62 & 3.95 $\pm$ 1.72 & 0.90 $\pm$ 0.07 & -- \\
        RoNIN-ResNet & 4.14 $\pm$ 2.00 & 2.38 $\pm$ 1.28 & 0.97 $\pm$ 0.03 & -- & 5.91 $\pm$ 3.24 & 3.79 $\pm$ 1.64 & 0.91 $\pm$ 0.07 & -- \\
        RoNIN-TCN & 4.75 $\pm$ 2.25 & 2.80 $\pm$ 1.33 & 1.27 $\pm$ 0.16 & 0.58 $\pm$ 0.57 & 5.57 $\pm$ 3.59 & 4.12 $\pm$ 2.05 & 1.25 $\pm$ 0.17 & 0.52 $\pm$ 0.60 \\
        RoNIN-LSTM & 4.47 $\pm$ 1.93 & 2.93 $\pm$ 1.41 & 1.27 $\pm$ 0.14 & 0.54 $\pm$ 0.58 & 6.18 $\pm$ 2.66 & 4.43 $\pm$ 2.11 & 1.24 $\pm$ 0.17 & 0.49 $\pm$ 0.60 \\
        TLIO & 3.78 $\pm$ 1.78 & 2.56 $\pm$ 1.26 & 0.95 $\pm$ 0.03 & -- & 4.90 $\pm$ 2.88 & 4.00 $\pm$ 1.87 & 0.89 $\pm$ 0.07 & -- \\
        LLIO & 4.66 $\pm$ 2.03 & 3.42 $\pm$ 1.50 & 0.92 $\pm$ 0.06 & -- & 6.14 $\pm$ 3.76 & 4.77 $\pm$ 2.38 & 0.84 $\pm$ 0.10 & -- \\
        \midrule
        \multicolumn{9}{l}{\textit{Ours}} \\
        PD single Domain & 2.23 $\pm$ 0.12 & 2.19 $\pm$ 0.25 & 0.66 $\pm$ 0.06 & 0.14 $\pm$ 0.61 & 2.25 $\pm$ 0.09 & 2.20 $\pm$ 0.22 & 0.64 $\pm$ 0.03 & 0.04 $\pm$ 0.61 \\
        PD Time Regr. & 3.48 $\pm$ 1.66 & 2.31 $\pm$ 1.20 & 0.97 $\pm$ 0.03 & 0.79 $\pm$ 0.44 & 5.14 $\pm$ 3.80 & \textbf{3.67 $\pm$ 1.80} & 0.91 $\pm$ 0.06 & \textbf{0.76 $\pm$ 0.47} \\
        PD Regression & 3.74 $\pm$ 1.56 & 2.42 $\pm$ 1.32 & 1.02 $\pm$ 0.04 & 0.79 $\pm$ 0.44 & 5.69 $\pm$ 4.06 & 3.91 $\pm$ 2.02 & \textbf{0.96 $\pm$ 0.09} & 0.75 $\pm$ 0.46 \\
        PD Time & 3.29 $\pm$ 1.41 & 2.25 $\pm$ 1.17 & 0.97 $\pm$ 0.03 & 0.79 $\pm$ 0.44 & 5.08 $\pm$ 2.55 & 3.69 $\pm$ 1.74 & 0.91 $\pm$ 0.06 & \textbf{0.76 $\pm$ 0.46} \\
        \textbf{PD (Ours)} & \textbf{3.17 $\pm$ 1.32} & \textbf{2.24 $\pm$ 1.23} & \textbf{0.99 $\pm$ 0.03} & \textbf{0.80 $\pm$ 0.44} & \textbf{4.82 $\pm$ 3.24} & 3.71 $\pm$ 1.76 & 0.93 $\pm$ 0.07 & \textbf{0.76 $\pm$ 0.46} \\
        \bottomrule
    \end{tabular*}
\end{table*}

%% file: table/comparison_SOTA.tex
\begin{table*}
    \centering
    \begin{threeparttable}
        \caption{\textbf{SOTA Quantitative Evaluation on RoNIN in 2D space.} Comparison against specialized 2D baselines on both seen and unseen environments. We project our PD from 3D predictions to 2D for fairness. Note that baselines only output velocity ($\vec{\mathbf{v}}$), whereas our method produces both velocity and angular velocity ($\vec{\mathbf{v}}, \boldsymbol{\vec{\omega}}$). Bold values indicate the best performance in each column.}
        \label{tab:sota_comparison}
        \begin{small}
            \setlength{\tabcolsep}{2.5pt}
            \begin{tabular*}{\linewidth}{l cc l c @{\extracolsep{\fill}} cccc cccc}
                \toprule
                & & & & & \multicolumn{4}{c}{\textbf{RoNIN (Seen)}} & \multicolumn{4}{c}{\textbf{RoNIN (Unseen)}} \\
                \cmidrule(lr){6-9} \cmidrule(lr){10-13}
                & & & & & \multicolumn{2}{c}{Position (m)} & \multicolumn{2}{c}{Orientation (rad)} & \multicolumn{2}{c}{Position (m)} & \multicolumn{2}{c}{Orientation (rad)} \\
                \cmidrule(lr){6-7} \cmidrule(lr){8-9} \cmidrule(lr){10-11} \cmidrule(lr){12-13}
                \textbf{Method} & \textbf{Out.} & \textbf{Dim.} & \textbf{Hardware} & \textbf{Time} & \textbf{ATE} $\downarrow$ & \textbf{RTE} $\downarrow$ & \textbf{ATE} $\downarrow$ & \textbf{RTE} $\downarrow$ & \textbf{ATE} $\downarrow$ & \textbf{RTE} $\downarrow$ & \textbf{ATE} $\downarrow$ & \textbf{RTE} $\downarrow$ \\
                \midrule
                CTIN~\cite{rao2022ctin}\tnote{*} & $\vec{\mathbf{v}}, \boldsymbol{\sigma}$ & 2D & RTX 2080 Ti & -- & 4.62 & 2.81 & -- & -- & 5.61 & 4.48 & -- & -- \\
                DiffusionIMU~\cite{teng2025diffusionimu}\tnote{*} & $\vec{\mathbf{v}}, \boldsymbol{\sigma}$ & 2D & NVIDIA A100 & 10 h & 3.64 & 2.72 & -- & -- & 5.27 & 4.31 & -- & -- \\
                EqNIO($O(2)$)~\cite{jayanth2025eqnio}\tnote{$\dagger$} & $\vec{\mathbf{v}}, \boldsymbol{\sigma}$ & 2D/3D & 2080 Ti & 1,4 h & 3.45 & 2.78  & -- & -- & \textbf{4.57} & 4.26 & -- & -- \\
                \midrule
                \textbf{PD (Ours)} & $\vec{\mathbf{v}}, \boldsymbol{\vec{\omega}}$ & 6D & 2$\times$ RTX 3090 & \textbf{10 h} & \textbf{3.15} & \textbf{2.23} & \textbf{1.78} & \textbf{1.55} & 4.79 & \textbf{3.70} & \textbf{1.83} & \textbf{1.59} \\

                \bottomrule
            \end{tabular*}
            \begin{tablenotes}
                \scriptsize
                \item[*] Results reported from original manuscript. We include their reported performance for reference, but direct comparisons should be made with caution due to potential differences in evaluation protocols and hardware capabilities.
                \item[$\dagger$] The result is replicate via the official codebase and weight, which is slightly different from the reported values, and the training time is reported by the authors.
            \end{tablenotes}
        \end{small}
    \end{threeparttable}
\end{table*}

%% file: table/performance_datasets.tex
\begin{table}
    \centering
    \caption{\textbf{Overall performance of the PedestrianDiffusion model.} We report performance based on $T=2$, where values are reported as mean $\pm$ standard deviation. \textbf{Bold} indicates the aggregated total performance.}
    \label{tab:PedestrianDiffusion_performance}
    \begin{small}
        \setlength{\tabcolsep}{2pt}
        \begin{tabular*}{\linewidth}{l @{\extracolsep{\fill}} c c c c}
            \toprule
            \textbf{Dataset} & \textbf{ATE} & \textbf{RTE} & \textbf{TLR} & \textbf{MCS} \\
            & $\downarrow$ & (1 min) $\downarrow$ & $\rightarrow 1$ & $\rightarrow 1$ \\
            \midrule
            \multicolumn{5}{l}{\textit{\textbf{Orientation (rad)}}}                                                                     \\
            \midrule
            OxIOD (V)$^{\dagger}$      & $0.59 \pm 0.48$          & $0.46 \pm 0.22$                  & $0.90 \pm 0.05$             & $0.86 \pm 0.28$ \\
            OxIOD (T)$^{\ddagger}$     & $1.54 \pm 0.34$          & $1.13 \pm 0.43$                  & $0.79 \pm 0.04$             & $0.69 \pm 0.43$ \\
            RIDI                       & $0.13 \pm 0.05$          & $0.17 \pm 0.09$                  & $0.99 \pm 0.00$             & $0.99 \pm 0.07$ \\
            RoNIN                      & $2.24 \pm 0.11$          & $2.14 \pm 0.19$                  & $0.88 \pm 0.03$             & $0.14 \pm 0.63$ \\
            RoNIN (U)$^{\mathsection}$ & $2.26 \pm 0.09$          & $2.15 \pm 0.23$                  & $0.86 \pm 0.02$             & $0.04 \pm 0.62$ \\
            TLIO                       & $0.12 \pm 0.06$          & $0.15 \pm 0.09$                  & $1.00 \pm 0.00$             & $0.99 \pm 0.04$ \\
            \midrule
            \textbf{Total}             & $\mathbf{1.26 \pm 1.05}$ & $\mathbf{1.21 \pm 0.99}$         & $\mathbf{0.92 \pm 0.07}$    & $\mathbf{0.35 \pm 0.67}$ \\
            \midrule
            \multicolumn{5}{l}{\textit{\textbf{Position (m)}}}                                                                         \\
            \midrule
            OxIOD (V)$^{\dagger}$      & $0.56 \pm 0.20$          & $0.30 \pm 0.03$                  & $0.97 \pm 0.02$             & $0.99 \pm 0.04$ \\
            OxIOD (T)$^{\ddagger}$     & $1.60 \pm 0.88$          & $1.34 \pm 0.73$                  & $1.03 \pm 0.02$             & $0.90 \pm 0.37$ \\
            RIDI                       & $1.38 \pm 0.79$          & $1.39 \pm 0.73$                  & $1.00 \pm 0.02$             & $0.96 \pm 0.17$ \\
            RoNIN                      & $3.02 \pm 1.40$          & $2.19 \pm 1.24$                  & $0.98 \pm 0.03$             & $0.80 \pm 0.44$ \\
            RoNIN (U)$^{\mathsection}$ & $4.79 \pm 2.79$          & $3.64 \pm 1.81$                  & $0.92 \pm 0.07$             & $0.76 \pm 0.46$ \\
            TLIO                       & $1.21 \pm 1.24$          & $0.98 \pm 2.01$                  & $0.98 \pm 0.07$             & $0.88 \pm 0.29$ \\
            \midrule
            \textbf{Total}             & $\mathbf{2.61 \pm 2.30}$ & $\mathbf{2.02 \pm 1.91}$         & $\mathbf{0.97 \pm 0.06}$    & $\mathbf{0.81 \pm 0.41}$ \\
            \bottomrule
            \multicolumn{5}{l}{\footnotesize $^{\dagger}$VICON subset \quad $^{\ddagger}$Tango subset \quad $^{\mathsection}$Unseen subset} \\
        \end{tabular*}
    \end{small}
\end{table}

%% file: table/computation.tex
\begin{table*}
    \centering
    \begin{threeparttable}
        \caption{\textbf{Computational Complexity and Efficiency Benchmarks.} Hardware performance evaluated using FP32 precision. By consolidating metrics, we contrast absolute processing costs against amortized per-frame efficiency, highlighting the generation density of our diffusion model over long sequential outputs ($N=100$).}
        \label{tab:computation}
        \small
        \renewcommand{\arraystretch}{1.15}
        \setlength{\tabcolsep}{4pt}
        \begin{tabular*}{\linewidth}{@{\extracolsep{\fill}} l ccccc cc cc cc @{}}
            \toprule
            & \multicolumn{5}{c}{\textbf{Model Specs}} & \multicolumn{2}{c}{\textbf{GPU Inf. (RTX 3090)}} & \multicolumn{2}{c}{\textbf{CPU Inf. (Edge)}} & \multicolumn{2}{c}{\textbf{Amort. Latency}} \\
            \cmidrule(lr){2-6} \cmidrule(lr){7-8} \cmidrule(lr){9-10} \cmidrule(lr){11-12}
            \textbf{Method} & \textbf{Output} & \textbf{Backbone} & \textbf{Param} & \textbf{FLOPs} & \textbf{$N$} & \textbf{VRAM} & \textbf{Lat.} & \textbf{RAM} & \textbf{Lat.} & \textbf{GPU} & \textbf{Edge} \\
             & & & (M) & (G) & (Fr) & (MB) & (ms) & (GB) & (ms) & (ms/fr) & (ms/fr) \\
            \midrule
            IoNet & $\vec{\mathbf{v}}, \boldsymbol{\omega}$ & ResNet & 1.16 & 0.40 & 1 & 300.48 & \textbf{3.54} & 2.37 & 405.08 & 3.54 & 405.08 \\
            RoNIN-ResNet & $\vec{\mathbf{v}}$ & ResNet-101 & 15.57 & 0.72 & 1 & 338.47 & 16.31 & 2.45 & 143.94 & 16.31 & 143.94 \\
            RoNIN-TCN & $\vec{\mathbf{v}}$ & TCN & \textbf{0.14} & 0.54 & 100 & \textbf{243.35} & 4.09 & 2.38 & \textbf{43.35} & \textbf{0.04} & \textbf{0.43} \\
            RoNIN-LSTM & $\vec{\mathbf{v}}$ & LSTM & 0.22 & \textbf{0.17} & 100 & 279.82 & 3.58 & 2.39 & 64.48 & \textbf{0.04} & 0.64 \\
            TLIO & $\vec{\mathbf{v}}, \boldsymbol{\sigma}$ & ResNet & 5.03 & 0.18 & 1 & 298.17 & 6.93 & 2.44 & 67.13 & 6.93 & 67.13 \\
            LLIO & $\vec{\mathbf{v}}, \boldsymbol{\sigma}$ & ResNet & 7.19 & 0.21 & 1 & 302.28 & 5.85 & 2.38 & 56.99 & 5.85 & 56.99 \\
            \midrule
            \textbf{PD (Ours)} & $\vec{\mathbf{v}}, \boldsymbol{\vec{\omega}}$ & UNet & 37.26\tnote{*} & 23.37 & \textbf{100} & 444.09 & 107.71 & \textbf{1.83} & 987.79 & 1.08 & 9.88 \\
            \bottomrule
        \end{tabular*}
        \begin{tablenotes}
            \scriptsize
            \item[*] Includes a 2.63M VAE encoder and a 34.63M Conditional UNet framework.
        \end{tablenotes}
    \end{threeparttable}
\end{table*}

%% file: table/ablation_clip_conditioning.tex
\begin{table*}
    \centering
    \caption{Comparison of performance results on the RIDI dataset ($T=1$). The rows distinguish between Unknown (without CLIP) and Known (with CLIP) configurations, while columns split the evaluation metrics (ATE, RTE, TLR, and MCS) for both Orientation and Position. All values are rounded to two decimal places, and superior performance results are highlighted in bold.}
    \label{tab:ablation_clip_conditioning}
    \begin{small}
        \setlength{\tabcolsep}{2pt}
        \begin{tabular*}{\textwidth}{l @{\extracolsep{\fill}} cccc cccc}
            \toprule
            \multirow{2}{*}{\textbf{Config.}} & \multicolumn{4}{c}{\textbf{Orientation (rad)}} & \multicolumn{4}{c}{\textbf{Position (m)}} \\
            \cmidrule(lr){2-5} \cmidrule(lr){6-9}
            & \textbf{ATE} & \textbf{RTE} & \textbf{TLR} & \textbf{MCS} & \textbf{ATE} & \textbf{RTE} & \textbf{TLR} & \textbf{MCS} \\
            \midrule
            \textbf{Unknown} & $0.29 \pm 0.34$ & $0.34 \pm 0.35$ & \boldmath$1.00 \pm 0.01$ & $0.99 \pm 0.10$ & $1.70 \pm 1.10$ & $1.60 \pm 0.80$ & $0.98 \pm 0.03$ & $0.95 \pm 0.19$ \\
            \textbf{Known}   & \boldmath$0.13 \pm 0.05$ & \boldmath$0.17 \pm 0.09$ & $0.99 \pm 0.00$ & \boldmath$0.99 \pm 0.07$ & \boldmath$1.38 \pm 0.79$ & \boldmath$1.39 \pm 0.73$ & \boldmath$1.00 \pm 0.02$ & \boldmath$0.96 \pm 0.17$ \\
            \bottomrule
        \end{tabular*}
    \end{small}
\end{table*}

%% file: Appendices.tex
\section{Appendix}
\input{table/datasets}
\subsection{Spectral-Domain Configuration}
\label{sec:appendix_input}
\input{./equation/global_standardization}
\autoref{tab:standardization_params} and \autoref{tab:Normalization} detail the hyperparameters for the spectral transformation and global standardization. We employ a hop size of 14 samples (\qty{0.14}{s} stride) to prioritize high-resolution trajectory reconstruction. However, these STFT parameters---particularly window overlap---can be adjusted to balance computational throughput against responsiveness, accommodating strict low-latency constraints during deployment, provided that hardware computational benchmarks permit it.

To formalize the uniform bound of the approximation error, $\epsilon$, across the highly anisotropic 6D kinematic manifold, we analyze the spectral preconditioning properties of the Short-Time Fourier Transform (STFT). In the raw temporal domain, the signal covariance $\Sigma_0$ is severely ill-conditioned due to vast scale and dynamic range disparities between translational velocity, $\vec{\mathbf{v}}$, and rotational velocity, $\boldsymbol{\vec{\omega}}$. The STFT projects these signals into a frequency-aligned basis, enabling modality-specific global standardization (normalized relative to $5\sigma_v$ and $5\sigma_\omega$). This diagonal scaling effectively preconditions the anisotropic covariance matrix, mapping the 6D manifold into an isotropic domain where the condition number is strictly bounded:
\begin{equation}
    \kappa(\Sigma_{\text{STFT}}) = \frac{\lambda_{\max}}{\lambda_{\min}} \leq C
\end{equation}
In the context of a single-step deterministic Probability Flow ODE solve (e.g., via DPM-Solver), the local truncation error is intrinsically bounded by the Lipschitz constant of the score estimator, $\nabla_{x_t}\log p_t(x_t)$. By enforcing a balanced signal-to-noise ratio (SNR) across both translational and rotational frequency bands, our spectral preconditioning averts asymmetric gradient inflation within the score network. This mathematical regularity guarantees that the approximation error $\epsilon$ remains uniformly and symmetrically bounded across majority kinematic dimensions.

\subsection{Datasets}
\label{sec:appendix_datasets}

We evaluate our method on four datasets: OxIOD~\cite{chen2018oxioddatasetdeepinertial}, RIDI~\cite{yan2017ridirobustimudouble}, RoNIN~\cite{9196860}, and TLIO~\cite{liu2020tlio}. These datasets employ diverse ground truth acquisition mechanisms, ranging from high-precision motion capture to smartphone-based SLAM. The OxIOD dataset contains a subset with VICON annotations, offering the highest precision in temporal and spatial alignment, albeit in small-scale controlled environments. For larger-scale environments, RIDI and the remainder of OxIOD utilize Google Tango, which leverages active depth sensing and visual-inertial SLAM. In contrast, RoNIN utilizes Google ARCore annotations, and the TLIO dataset relies on a proprietary VIO pipeline. TLIO is notably the largest dataset, comprising 60 hours of data with complex 3D motion (e.g., stair traversal), whereas the other datasets primarily feature 2D planar trajectories. This diverse combination of high-precision and large-scale data ensures comprehensive coverage of real-world scenarios, enhancing the generalizability of our model. Detailed dataset statistics are provided in \autoref{tab:dataset_summary}.

%% file: table/datasets.tex
\begin{table*}
    \centering
    \caption{Detailed statistics for the OxIOD, RIDI, RoNIN, and TLIO datasets. \textbf{Seq}: Number of sequences, \textbf{Samples}: Number of samples (windows). Percentages in parentheses represent the contribution of each dataset to the combined hybrid dataset.}
    \label{tab:dataset_summary}
    \begin{small}
        \resizebox{\linewidth}{!}{
            \begin{tabular}{l c rr rr rr}
                \toprule
                \multirow{2}{*}{\textbf{Dataset}} & \multirow{2}{*}{\textbf{GT}} & \multicolumn{2}{c}{\textbf{Train}} & \multicolumn{2}{c}{\textbf{Validation}} & \multicolumn{2}{c}{\textbf{Test}}                                                                                       \\
                \cmidrule(lr){3-4} \cmidrule(lr){5-6} \cmidrule(lr){7-8}
                                                  &                              & \textbf{Seq}                       & \textbf{Samples}                        & \textbf{Seq}                      & \textbf{Samples}                 & \textbf{Seq} & \textbf{Samples}                  \\
                \midrule
                OxIOD (VICON)                     & VICON                        & 138 (19.8\%)                       & 123,874 (12.8\%)                        & 22 (21.0\%)                       & 20,514 (14.2\%)                  & 6 (4.6\%)    & 581 (1.2\%)                       \\
                OxIOD (Tango)                     & Tango                        & 138 (19.8\%)                       & 12,153 (1.3\%)                          & 22 (21.0\%)                       & 736 (0.5\%)                      & 6 (4.6\%)    & 303 (0.6\%)                       \\
                RIDI                              & Tango                        & 66 (9.5\%)                         & 39,514 (4.1\%)                          & 10 (9.5\%)                        & 5,952 (4.1\%)                    & 18 (13.8\%)  & 2,174 (4.3\%)                     \\
                \multirow{2}{*}{RoNIN}            & \multirow{2}{*}{ARCore}      & \multirow{2}{*}{72 (10.3\%)}       & \multirow{2}{*}{240,838 (24.9\%)}       & \multirow{2}{*}{16 (15.2\%)}      & \multirow{2}{*}{49,348 (34.1\%)} & 32 (24.6\%)  & 17,713 (35.2\%) \textit{(Seen)}   \\
                                                  &                              &                                    &                                         &                                   &                                  & 32 (24.6\%)  & 18,218 (36.2\%) \textit{(Unseen)} \\
                TLIO                              & VIO                          & 283 (40.6\%)                       & 549,814 (56.9\%)                        & 35 (33.3\%)                       & 67,969 (47.0\%)                  & 36 (27.7\%)  & 11,379 (22.6\%)                   \\
                \bottomrule
            \end{tabular}%
        }
    \end{small}
\end{table*}

%% file: equation/global_standardization.tex
\begin{table}
    \centering
    \caption{Hyperparameters employed for the spectral transformation and global normalization strategy.}
    \label{tab:standardization_params}
    \small
    \begin{tabular*}{\linewidth}{l @{\extracolsep{\fill}} c}
        \toprule
        \textbf{Parameter} & \textbf{Value} \\
        \midrule
        Sampling Rate ($f_s$) & $\approx 100$\,Hz \\
        STFT Window Size & 30 \\
        FFT Points ($N_{\text{fft}}$) & 30 \\
        Hop Size & 14 \\
        Spectral Output Shape & $(16, 8, 2)$ \\
        Normalization Method & Global Max Std \\
        \bottomrule
    \end{tabular*}
\end{table}
\begin{table}
    \centering
    \caption{Normalization Parameters (Spectral Domain)}
    \label{tab:Normalization}
    \small
    \sisetup{table-format=2.2}
    \begin{tabular*}{\linewidth}{l @{\extracolsep{\fill}} S l}
        \toprule
        \textbf{Signal Component} & {\textbf{Std. Dev. ($5 \times \sigma$)}} & \textbf{Physical Unit} \\
        \midrule
        Spectral Accel.       & 28.91 & $\mathrm{m/s^2}$ \\
        Spectral Gyro.        & 11.92 & $\mathrm{rad/s}$ \\
        Spectral Velocity     & 65.76 & $\mathrm{m/s}$ \\
        Spectral Ang. Vel.    & 14.01 & $\mathrm{rad/s}$ \\
        \bottomrule
    \end{tabular*}
    \vspace{0.5em}
    \begin{minipage}{\linewidth}
        \footnotesize{\emph{Note: Values represent the scaled standard deviation of the Short-Time Fourier Transform (STFT) magnitudes $\times 5$.}}
    \end{minipage}
\end{table}